\documentclass[12pt]{article}

\newcommand{\Waux}{W^{\mathrm{aux.}}}

\newif\ificml
\newif\ifproofsinbody
\newif\ifwatermark

\ificml
    \usepackage{icml2025}
\else
    \usepackage[top=1in, bottom=1in, left=1in, right=1in]{geometry}
    \usepackage{algorithm}
    \usepackage{algorithmic}
    \usepackage{titlesec}
    \usepackage{fancyhdr}
    \usepackage{authblk}

\fi
\usepackage{tikz}
\usepackage{xurl}
\usepackage{microtype}

\usepackage{setspace}
\usepackage[shortlabels]{enumitem}
\usepackage{booktabs}
\usepackage{rotating}
\usepackage{graphicx}
\usepackage[section]{placeins} 
\usepackage[bf]{caption}
\usepackage{subcaption}
\ificml
\else
    \usepackage{xcolor}
\fi
\usepackage{bbm}

\usepackage{natbib}

\bibpunct{(}{)}{;}{a}{,}{,}

\usepackage{amsmath, amsfonts, amssymb, amsthm}
\allowdisplaybreaks

\usepackage[T1]{fontenc}
\usepackage{svg}
\svgsetup{
  inkscapelatex=false,  
}

\ificml
\usepackage[pdftex]{hyperref}
\else

\usepackage[colorlinks=true, linkcolor=black, citecolor=blue, urlcolor=black, filecolor=blue, pdfborder={0 0 0}]{hyperref}
\fi

\usepackage{mathtools}
\usepackage[capitalize,noabbrev]{cleveref}

\theoremstyle{plain}
\newtheorem{theorem}{Theorem}[section]

\newtheorem{lemma}[theorem]{Lemma}

\theoremstyle{definition}
\newtheorem{definition}[theorem]{Definition}
\newtheorem{assumption}[theorem]{Assumption}
\theoremstyle{remark}

\newcommand{\E}{\mathbb{E}}

\newcommand{\abs}[1]{\left| #1 \right|}

\newcommand{\Var}{\mathrm{Var}}
\newcommand{\Corr}{\mathrm{Corr}}
\newcommand{\Cov}{\mathrm{Cov}}
\newcommand{\convinprob}{\overset{p}{\to}}
\newcommand{\convindist}{\overset{d}{\to}}

\numberwithin{equation}{section}

    \newcommand{\plugin}{\varphi^{\mathrm{pi}}}
    \newcommand{\debiasing}{\varphi^{\mathrm{db}}}

\newcommand{\cB}{\mathcal{B}}

\newcommand{\cD}{\mathcal{D}}
\newcommand{\cE}{\mathcal{E}}
\newcommand{\cF}{\mathcal{F}}

\newcommand{\cN}{\mathcal{N}}
\newcommand{\cP}{\mathcal{P}}
\newcommand{\cS}{\mathcal{S}}

\newcommand{\cW}{\mathcal{W}}

\newcommand{\indep}{\perp \!\!\!\perp}

\newcommand{\defeq}{\overset{\mathrm{def.}}{=}}

\newcommand{\indic}[1]{\mathbbm{1}\left\{{#1} \right\}}

\newcommand{\R}{\mathbb{R}}

\newcommand{\N}{\mathbb{N}}

\newcommand{\prob}[1]{\mathbb{P}\left[ {#1} \right]}
\newcommand{\Unif}{\mathrm{Unif}}

\newcommand{\floor}[1]{\lfloor {#1} \rfloor}
\newcommand{\paren}[1]{\left( {#1} \right)}
\newcommand{\norm}[1]{\left\| {#1} \right\|}
\newcommand{\tvnorm}[1]{\left\| {#1} \right\|_{\mathrm{TV}}}

\newcommand{\sqparen}[1]{\left[ {#1} \right]}
\newcommand{\ex}[1]{\mathbb{E}\sqparen{#1}}
\newcommand{\expt}[1]{\E_{P}\sqparen{#1}}

\newcommand{\curly}[1]{\left\{ {#1} \right\}}

\newcommand{\probgiv}[2]{P\left( {#1} \;\middle|\; {#2} \right)}

\newcommand{\ch}[1]{\textcolor{olive}{[CH: {#1}]}}
\newcommand{\mr}[1]{\textcolor{red}{[MR: {#1}]}}

\renewcommand{\ch}[1]{}
\renewcommand{\mr}[1]{}
\newcommand{\ditto}{---\texttt{"}---}

\crefname{assumption}{Assumption}{Assumptions}
\Crefname{assumption}{Assumption}{Assumptions}

\newcommand{\1}{\mathbf{1}}
\newcommand{\0}{\mathbf{0}}

\crefname{enumi}{}{items}
\Crefname{enumi}{}{Items}

\newcommand{\dvec}{D_{(t-m):t}}
\newcommand{\xvec}{X_{(t-m):t}}

\newcommand{\textvec}{\mathtt{vec}}
\newcommand{\Ber}{\mathrm{Ber}}

\ificml
    \ifwatermark
        \AtBeginDocument{%
            \AddToShipoutPicture{%
                \setlength{\unitlength}{1pt}%
                \put(100,750){%
                    \color[gray]{0.8}%
                    \textbf{DRAFT: PLEASE DO NOT DISTRIBUTE}%
                }%
            }%
        }
    \fi
\else
    \title{Double Machine Learning for Causal Inference under Shared-State Interference}
    \author{Chris Hays}
    \author{Manish Raghavan}
    \affil{MIT}
    \affil{\texttt{\{jhays,mragh\}@mit.edu}}

    \ifwatermark
    \fancypagestyle{plain}{
        \fancyhf{} 
        \fancyhead[C]{\textcolor{gray}{WORKING DRAFT}}
        \fancyfoot[C]{\thepage}
        
    }
    \fi
\fi

\begin{document}

\ificml
\else
    \maketitle
\fi


\begin{abstract}
    Researchers and practitioners often wish to measure treatment effects in settings where units interact via markets and recommendation systems.  In these settings, units are affected by certain \textit{shared states}, like prices, algorithmic recommendations or social signals. We formalize this  structure, calling it {shared-state interference}, and argue that our formulation captures many relevant applied settings. Our key modeling assumption is that individuals' potential outcomes are independent conditional on the shared state. We then prove an extension of a double machine learning (DML) theorem providing conditions for achieving efficient inference under shared-state interference. We also instantiate our general theorem in several models of interest where it is possible to efficiently estimate the average direct effect (ADE) or global average treatment effect (GATE).
\end{abstract}

\section{Introduction}

    In causal inference, interference --- where individuals' treatment assignments, outcomes or other characteristics impact others' outcomes --- is everywhere:
    On short-form video platforms, the consumption of one individual is used as an input to recommendation algorithms that are subsequently used to serve content to others.
    %
    In ride-sharing or housing rental services, providing a discount to some people can cause them to increase their usage of the service and increase wait times or prices.
    %
    %
    In such settings, failure to account for interference may lead to biased estimates of causal effects, even in randomized controlled trials.

    Across many settings, like those of recommender systems and markets, interference between units often follow a common pattern: the outcomes of individuals depend on others through some \textit{shared state}.
    %
    In recommender systems, the shared state might be the outputs of recommender systems that are used to generate users' feeds.
    In marketplaces, the shared state may be prices, wait times, or availability of inventory.
    In each of these settings, individuals both influence and are influenced by the shared state.
    %

    Assumptions are required to perform causal inference in the presence of interference.
    Prior work has studied this kind of interference in particular applications \citep{johari_experimental_2021,basse_randomization_2016,wager_causal_2024,munro_causal_2024,simchi-levi_pricing_2023,dhaouadi_price_2023,farias_correcting_2023,farias_markovian_2022,li_experimenting_2024}.
    In some cases, they propose either experimental designs or analysis methods that account for interference and allow for valid inference.
    However, these approaches typically assume a parametric model of the system or domain-specific assumptions.
    %


    \paragraph{Our contributions.}
    We provide a general framework to reason about causal inference under shared-state interference. We define a formal model in \Cref{sec:model}, where individuals arrive sequentially. Their covariates (i.e., unit characteristics), treatment assignments, and outcomes may influence the outcomes of future individuals through some shared state.
    
    In \Cref{sec:dml4ssi}, we provide conditions under which efficient inference is possible in the framework.
    We extend methods from double machine learning (DML) \citep{chernozhukov_doubledebiased_2018}, which has shown that it is possible to achieve efficient inference using expressive machine learning methods without parametric functional form assumptions.
    A key assumption we rely on is that the shared state progresses according to a Markov chain with particular properties.
    We also provide a consistent variance estimator, which is necessary for constructing valid confidence intervals and running hypothesis tests.

    Next, we instantiate our framework with two applications: first, we show how our framework can be used to estimate average direct effects (ADE) in observational settings; then, we provide a variance reduction strategy for estimation of the global average treatment effects (GATE) in switchback experiments.
    %
    %
    Intuitively, the ADE measures the expected difference between treatment and control for a unit drawn uniformly at random, keeping the treatment assignment distributions of other units fixed.
    The GATE measures the mean difference between outcomes when all units are assigned to treatment versus when all are assigned to control.
    Key to our approach in these sections is constructing structural models that capture settings of interest and estimators for the treatment effects which satisfy the properties necessary to apply the DML theorem from \cref{sec:dml4ssi}.
    In each instantiation, we provide simulations validating that our method produces estimators that concentrate around the true treatment effects in finite samples. We also show our consistent variance estimators can be used to construct confidence intervals with the desired coverage probability.

\subsection{Related work.} 
Our work sits at the intersection of the study of inference under interference and double machine learning (DML) methods.

\paragraph{Shared-state interference.} Several prior works have explored causal inference in settings captured by or similar to our formulation of shared-state interference \citep{johari_experimental_2021,basse_randomization_2016,wager_causal_2024,munro_causal_2024,holtz_reducing_2024,simchi-levi_pricing_2023,dhaouadi_price_2023,farias_correcting_2023,farias_markovian_2022,li_experimenting_2024,brennan_reducing_2024,bajari_experimental_2023,bright_reducing_2023}.
Some of these papers have analyzed bias resulting from naive estimators or experimental designs and proposed less biased alternatives \citep{johari_experimental_2021, farias_markovian_2022,brennan_reducing_2024}.
Other works have proposed parametric models of settings that allow for inference in the presence of interference \citep{wager_experimenting_2021, li_experimenting_2024}.
Several works consider a Markov model of interference similar to ours in an experimental setting and explore less biased estimation methods \citep{farias_correcting_2023,farias_markovian_2022,glynn_adaptive_2022}.

\paragraph{Double machine learning.}
The double machine learning framework was introduced in \citet{chernozhukov_doubledebiased_2018}, drawing on a rich literature in semiparametric statistics (see, e.g., \citet{kennedy_semiparametric_2023} for an overview).
%
DML methods have been extended to several other settings, such as those with continuous treatments \citep{kennedy_non-parametric_2017}, estimation of quantile effects \citep{kallus_localized_2022} and settings with limited unobserved confounding \citep{rambachan_robust_2024}.
A few works, like ours, have developed semiparametric methods for settings with interference. 
Several works have explored interference channeled through individuals' social networks \citep{emmenegger_treatment_2023,ogburn_causal_2024}.
Our setting is different because each unit may be dependent on each of the units before it (whereas their network-based approaches require the network to be sparse), and interference in our setting is anonymous (i.e., independent of the identities of the units).
%
\citet{munro_causal_2024} develops a DML theorem for settings like auctions where interference is channeled through a centralized allocation mechanism.
%
\citet{zhan_estimating_2024} develops a DML theorem for a discrete choice model of content consumption under a neural network-based recommender system.
%
In contrast, our work relaxes these assumptions and instead requires a general Markovian assumption.
%
\citet{ballinari_semiparametric_2024} develops a DML theorem for a time-series model where a single unit is observed over time, but where observations across time obey a mixing condition.
Their method allows for estimating impulse response functions, which measure the effect of an intervention at a particular time on the outcome at a future time.
Our work also relies on mixing conditions for our convergence results, but our model assumes all dependence between observations is channeled through the shared-state, and our model instantiations and estimands in \cref{sec:ade,sec:sb} are different from theirs.
We supply an extended comparison to some key related works in \cref{sec:morerelatedwork}.

%
%
%

%

\section{Modeling Shared-State Interference} \label{sec:model}

We first give a high-level description of our setting. 
Our model begins with a set of sequentially arriving units.
%
Each unit has covariates and a binary treatment assignment drawn iid from a joint distribution, as in canonical causal inference settings.
There is also a (possibly vector-valued) observed shared state through which all interference is channeled: each arriving unit has an outcome of interest that may depend on the shared state, their covariates and their treatment assignment; then the shared state at the next time step may depend on the previous shared state, as well as the previous unit's outcome, covariates and treatment assignment. 
The key assumption that makes our setup tractable is that the shared state has a Markov property: that is, conditional on the shared state at the previous time step and the data of the unit that arrived in the previous time step, the shared state is independent of the previous data.
We next introduce the model and notation formally.
%

\subsection{Setting}

    \paragraph{Notation.} Capital letters will denote random variables. For a vector ${x}$ and indices $i,j$, ${x}_{i:j}$ indicates the column vector of entries $(x_{i}, \dots, x_{j})'$ where $'$ indicates the transpose. For a constant $c$, the vector  $(c, c, \dots, c)'$ (of length induced by context) will be denoted using bold font $\mathbf{c}$.  
    For a scalar random variable $V \sim P$, we will denote the $L^q(P)$ norm $\norm{V}_{L^q(P)} = \E_P[{\abs{V}^q}]^{1/q}$.
    For two functions $f$ and $g$ taking the same argument $x$, we may write $(f+g)(x) = f(x) + g(x)$ to denote their sum.

    \paragraph{Model.} We will consider a sequence of units indexed $t=1,2,\dots,$ and observed up to time $T$. 
    Each unit $t$ will have features $X_t$ drawn iid from a sample space $\R^{p_X}$ for finite dimension $p_X$ and a binary treatment assignment $D_t$.
    We will let $H_t$ on $\R^{p_H}$, for finite dimension $p_H$, denote the shared state.
    The observed outcome for each unit will be real-valued and denoted $Y_t$.
    
    We will adopt notation so that potential outcomes are a (stochastic) function of treatment and the shared-state (i.e., $Y_t(D_t, H_t)$) similar to the exposure mapping approach of \citet{aronow_estimating_2017}.\footnote{Exposure mappings are typically defined to be functions of the set of \textit{treatment assignments}, but in our setting, as we will discuss, $H_t$ may depend on other units' (stochastic) \textit{outcomes}.}
    The dependence of outcomes on the shared state is the primary difference between this model and canonical iid causal inference settings, and the avenue through which interference is assumed to occur.
    We will assume $D_t$ is independent of all data from prior units but that it may depend on $X_t$.
    The observed data associated with time step $t$ will be denoted $W_t = \{ X_t, D_t, H_t, Y_t \}$ on the sample space $\cW = \R^{p_X } \times \{ 0, 1\} \times \R^{ p_H} \times \R$. We will denote the full observed data $W_{1:T} = \{ W_t \}_{t=1}^T \in \cW^T$ and the (unknown, joint) data distribution of $W_{1:\infty} = \{W_t\}_{t=1}^\infty$ as $P \in \cP$ for some set of distributions $\cP$.
    The set $\cP$ specifies the set of probability distributions in which the true distribution may fall, and we will impose assumptions that must hold for all $P \in \cP$.
    %

    %
    %
    %
    %
    
    We will assume that $H_t$ depends on the data from the previous time step (i.e., $W_{t-1}$) but is independent of $X_t$, and is conditionally independent of the data at time steps before time $t-1$, i.e., that $H_t$ has the Markov property: For all $P \in \cP$ and $P$-measurable $A \subseteq \R^{p_H}$, and $t = 1, 2, \dots$
    \begin{align*}
        P(H_t \in A \; | \; W_{1:(t-1)}) = P(H_t \in A \; | \; W_{t-1}) 
    \end{align*}
    This is assumption satisfied in settings where $H_t$ is an update to $H_{t-1}$, like when inventory at time $t$ is a function of inventory at time $t-1$ and whether the customer at time $t-1$ made a purchase.
    
\begin{figure}
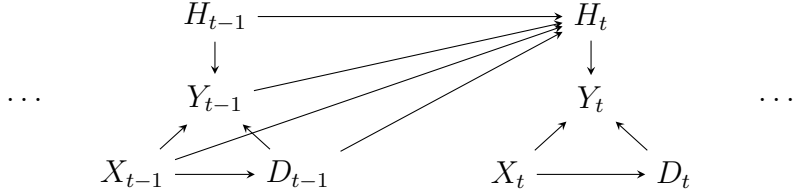

        \centering
        \tikzset{
            > = stealth,
            every node/.append style = {
                text = black
            },
            every path/.append style = {
                arrows = ->,
                draw = black,
            },
            hidden/.style = {
                draw = black,
                shape = circle,
                inner sep = 1pt
            }
        }
        \tikz{
            \node (d0) at (0,2) {$\dots$};
            
            \node (h3) at (10,2) {$\dots$};
            
            \node (x1) at (1.4,1) {$X_{t-1}$};
            \node (d1) at (3.6,1) {$D_{t-1}$};
            \node (y1) at (2.5,2) {$Y_{t-1}$};
            \node (h1) at (2.5,3.1) {$H_{t-1}$};
        
            \node (x2) at (6.4,1) {$X_t$};
            \node (d2) at (8.6,1) {$D_t$};
            \node (y2) at (7.5,2) {$Y_t$};
            \node (h2) at (7.5,3.1) {$H_t$};

            \path (x1) edge (d1);
            \path (d1) edge (y1);
            \path (h1) edge (y1);
            \path (x1) edge (y1);
        
            \path (x2) edge (d2);
            \path (d2) edge (y2);
            \path (h2) edge (y2);
            \path (x2) edge (y2);

            \path (x1) edge (h2);
            \path (d1) edge (h2);
            \path (h1) edge (h2);
            \path (y1) edge (h2);

        }
        \caption{Dependency structure of shared-state interference.}
        \label{fig:dag}
    \end{figure}

    Our setting allows for the shared state at time $t$ to depend on the shared state at time $t-1$, allowing for dependencies over time, as long as these dependencies are exclusively mediated by the shared state.
    For example, the purchasing decisions of two individuals may be correlated if inventory is low during the time interval in which they arrive.
    The shared-state at time $t$ can also depend on the \textit{outcome} at time $t-1$. 
    This is different from many other models for causal inference under interference, which allow for dependencies only on other units' \textit{treatment assignments}.
    For example, the canonical neighborhood treatment response assumption \citep{eckles_design_2014}, informally, says that, for a particular network, a node's potential outcomes depend only on its treatment assignment and its neighbors' treatment assignments.
    Thus, our method has the desirable property that individuals' outcomes can depend on prior individuals' \textit{behavior}, not just prior treatment assignments.
    The dependency structure assumed by a shared-state interference setting is summarized in \cref{fig:dag}.

    We will also assume that the distribution of $H_t$ and $D_t$ conditional on $W_{t-1}$ is invariant of $t$. Thus, we can specify a transition probability kernel $K_H$ where, for $w \in \cW$ and measurable event $A$ on $\cW$ and all $t \in [T]$,
    \begin{align}
        K_H(w, A) \defeq \probgiv{H_{t} \in A}{W_{t-1} = w}.
    \end{align}
    There will be an initial (known) shared-state $H_0$ generated according to an arbitrary, unknown and possibly deterministic distribution $P_{0}$.
    The $t$-step kernel will be denoted $K_H^t$, where $K_H^t(w, A) = P(W_{s+t} \in A \; | \; W_{s} = w)$.

    The fact that $H_t$ has the Markov property, $X_t$ is drawn iid, and $D_t$ is independent of the prior data trivially implies that $\{ W_t\}_{t=1}^\infty$ itself is a Markov chain.
    Indeed, since all transition probabilities are $t$-invariant, $\{ W_t\}_{t=1}^\infty$ is homogeneous.
    $\cW$ may be uncountable, so the chain may be defined on a general (not necessarily finite) state space.
    We will denote the (unknown) transition probability kernel of $W_{1:\infty}$ as $K$ where $K(w, A) = \probgiv{W_t \in A}{W_{t-1} = w}$. Note that $K$ is distinct from $K_H$; the former denotes the transition probabilities for $W_t$ and the latter only for $H_t$. As with $K_H$, the $t$-step kernel will be denoted $K^t$.
    
    When the Markov chain has a unique stationary distribution, we will denote it $K^\infty$, since we will only be considering settings where, for all $w, A$ it holds $K^\infty(A) = \lim_{T \to \infty} K^T(w, A)$.
    (We drop the argument $w$ in $K^\infty$ since $K^\infty$ does not depend on the starting state.)
    When $K^\infty$ exists, we will require $K^\infty \in \cP$.

    We will also require that $\{ W_t \}_{t=1}^\infty$ satisfies natural Markov chain conditions.

    \begin{assumption} \label{assm:markovchain}
        $W_{1:\infty}$ satisfies at least one of the following conditions: 
        \begin{enumerate}[(a)]
            \item \label{item:geoerg}Geometric ergodicity and detailed balance: i.e., $W_{1:\infty}$ is a Harris ergodic Markov chain satisfying geometric mixing, and for $w, z \in \cW$, it holds
            \begin{align}
                K^\infty(dw) K(w, dz) = K^\infty(dz) K(z, dw). \label{def:detailedbalance}
            \end{align}
            \item \label{item:mdep} $m$-dependence: i.e., 
            \begin{align}
                W_1, \dots, W_{t} \indep W_{t+m+1}, \dots, W_T \label{def:mdep}
            \end{align}
            for all $t \in [T]$. 
        \end{enumerate}
    \end{assumption}
    Geometric ergodicity is a generalization (to the general state space setting) of aperiodicity and irreducibility in finite-state Markov chains; 
    we define it formally in \cref{sec:background}, and refer the reader to standard references like \citet{meyn_markov_2009} 
    for a complete treatment of general state space Markov chains.
    \cref{assm:markovchain}\cref{item:geoerg} implies that the chain has a unique steady state transition probability distribution and that the chain converges to that distribution from any starting state \citep[Proposition 17.1.6]{meyn_markov_2009}. Thus, writing $K^\infty$ in \cref{def:detailedbalance} is well-defined.
    Intuitively, detailed balance says that, for continuous distributions, the steady state probability density around a point $w$ times the probability density of transitioning from $w$ to $z$ is equal to the same expression with the roles of $w,z$ reversed.
    %
    %
    
    The $m$-dependence condition states that data observed more than $m$ steps apart are independent.
    Observe that any $m$-dependent sequence can be written as a Markov chain, using the last $m$ observations as the state.
    We will assume throughout when invoking \cref{assm:markovchain}\cref{item:mdep} that (a finite upper bound on) $m$ is known.
    We refer the reader to the literature on switchback experiments for methods of estimating $m$ from data \citep{bojinov_design_2022}.

    \paragraph{Estimand.} We will assume there is a scalar-valued functional of interest $\psi^*$ which depends on the data distribution $P$; e.g., in \cref{sec:ade}, $\psi^*$ will be the average direct effect and in \cref{sec:sb}, it will be the global average treatment effect. We will assume there exists a function $\varphi^* \; : \; \cW \to \R$ so that $\psi^*$ can be written as an asymptotic average over expected values of $\varphi^*$.
    I.e.,
    \begin{align*}
        \psi^* = \lim_{T \to \infty} \frac{1}{T} \sum_{t=1}^T\expt{\varphi^*(W_t)}.
    \end{align*}
    We will assume that $\varphi^*$ is continuous.



\paragraph{Examples of shared-state interference.} \label{sec:instantiate}

    %

    
    Our model is suited to describing marketplaces where interference is induced by limited supply or demand. For example, \citet{johari_experimental_2021} proposes a Markov chain model of a rental platform consisting of a sequence of arriving customers and a set of listing types, each with a fixed number of available listings. 
    If a customer books a listing, it becomes unavailable for a time, temporarily decreasing the number of listings of the type by 1.
    %
    %
    %
    Several other models of interference in specific markets can be well-captured by shared-state interference: for example, models of freelance labor markets \citep{wager_experimenting_2021} and ride-sharing markets \citep{li_experimenting_2024}.
    Moreover, $m$-dependence is assumed in the large literature on switchback experiments \citep{bojinov_design_2022}.

    %
    %

\section{DML for SSI} \label{sec:dml4ssi}

    In this section, our main result is a double machine learning (DML) theorem for causal inference under shared state interference (SSI).
    Double machine learning is a meta-algorithm which allows for efficient inference with the use of expressive machine learning methods. 
    It was proposed in \citet{chernozhukov_doubledebiased_2018}, building on a large literature in semiparametric statistics.
    We provide background on the method in \cref{sec:background} and give an informal overview here. 
    We refer the reader to \citet{chernozhukov_doubledebiased_2018} for a full treatment of the topic in the iid setting and to \citet{chernozhukov_applied_2024} for a gentle introduction.
    
    At a high level, the method consists of two steps.
    First, an expressive machine learning algorithm is used to approximate any nuisance parameters.
    Nuisance parameters often take the form of estimated conditional expectation functions (the expected outcome conditional on covariates and treatment) or propensity scores (probability of treatment conditional on covariates) which are necessary for estimation of treatment effects but not of interest on their own. 
    A predictor for the expected outcome conditional on covariates could be used to create plug-in estimators by simply taking the average of predicted outcomes under treatment versus under control.
    However, such plug-in estimators may converge at slower-than-parametric rates and may be very biased in finite samples.
    Thus, second, the estimator is constructed so as to satisfy a first-order insensitivity property, in a distributional sense that we will precisely specify later.
    This property can be used to show that the convergence rates of the estimator depend only on \textit{products} of convergence rates of the ML estimators, which allows for weaker conditions on the convergence rates for any one estimator.
    
    We note that a naive approach to inference in our setting would be to treat the shared-state as a covariate and apply DML methods as if the data were iid.
    However, depending on the estimand, this may yield inconsistent treatment effect or variance estimators, as we will see in our simulations in \cref{sec:ade,sec:sb}.

    An important part of any DML procedure is that the nuisance parameter estimators must be independent of the data used to construct the target estimator.
    (Otherwise, typical $L^2$ convergence rates are not sufficient to ensure that the nuisance estimators yield consistent target estimators.)
    In iid settings, this independence is usually achieved via sample splitting: the data is split into $k$ folds, the ML estimator is trained on all but one fold and then the estimator is constructed using the held-out fold.
    In our setting, in the vein of \citet{angelopoulos_prediction-powered_2023}, \citet{ballinari_semiparametric_2024} and many similar methods in statistics and machine learning, we instead assume that the nuisance estimators are generated via an auxiliary sample of data that is independent of the data to be used for inference.
    We make this simple alternate assumption because sample splitting would not guarantee independence between folds of the data in our setting, so predictors trained on some folds may still not be independent of the others.
     
    If the data is $m$-dependent, then sample splitting can be used to generate a predictor that is independent of the data used to construct the estimator.
    Additionally, if the structural equation models assumed about the data generating process are Donsker,\footnote{Donsker classes, intuitively, are function classes that are simple enough that they cannot arbitrarily overfit on training data.
    Vapnik-Chervonenkis (VC) classes and bounded monotone function classes, among many others, are Donsker.
    See \citet{kennedy_semiparametric_2016} for detailed discussion of this approach.} no auxiliary sample is necessary.
    Finally, we note that in our interference setting, units at far apart time steps --- even without $m$-dependence --- have approximately uncorrelated outcomes, so it may be possible to split the data into approximately uncorrelated subsamples.
    We leave for future work the exploration of whether predictors trained on approximately uncorrelated data splits can be used for valid inference.
    
    %
    For our theorem in this section, we will assume that there is a known estimator $\psi$ that depends on $W_{1:T}$ and a set of data-dependent nuisance parameters/functions $\eta \in \cS$ for some convex set $\cS$.
    We can imagine $\eta$ to be the parameters in a neural network, the coefficients on a possibly high-dimensional linear regression, or the feature splits and thresholds for trees in a random forest. 
    We will need that $\psi$ identifies $\psi^*$; i.e., that $\expt{\psi(W_{1:T}; \eta^*)} = \psi^*(P)$ for unknown true nuissances  $\eta^* \in \cS$.
    We will also require that $\psi$ is an empirical mean of a time-$t$ estimator $\varphi$, i.e., that
    \begin{align*}
        \psi(W_{1:T}; \eta) = T^{-1} \sum_{t=1}^T \varphi(W_t; \eta).
    \end{align*}
    In \Cref{sec:ade,sec:sb}, $\psi$ will be an estimator for the average direct effect (ADE) and global average treatment effect (GATE), respectively, which we define formally under the structural models in each of those sections.
    The estimator will be specially constructed so as to satisfy the double-robustness property characteristic of DML methods.
    Much of the challenge in applying this section's main theorem involves specifying a structural model that fits a context of interest and identifying an estimator with the double-robustness properties.

    We will assume access to a nuisance function estimator $\hat \eta$ (usually imagined to have been trained by a flexible machine learning method like a neural network or random forest).
    We will also require that $\hat \eta$ belongs to a convex set containing $\eta^*$, with high probability.
    This is for technical reasons that will become clear later and are common to all DML methods: the conditions required for our main theorem necessitate reasoning over convex combinations of $\hat \eta$ and $\eta^*$ so it is useful to define a convex set within which $\hat \eta$ and $\eta^*$ fall.
    Finally, we will require that the convex set is shrinking around $\eta^*$ asymptotically in $T$.
    This ensures that $\hat \eta$ is converging to $\eta^*$ (in an appropriate sense) as $T$ scales.
    Formally, for a constant $0 \leq \gamma < 1$, we will require that the nuisance function estimator $\hat \eta$ belongs to a convex set $S_T \subseteq \cS$ with probability $1 - \gamma$ where it is also assumed $\eta^* \in S_T$.
    The DML procedure for our context is summarized in \cref{alg:dmlssv}.
    
    \begin{algorithm}[t]
    \caption{Double machine learning for shared-state interference}\label{alg:dmlssv}
        \begin{algorithmic}
        \STATE Train a nuisance function estimator $\hat \eta$ from auxiliary data $\Waux$.
        \STATE Construct the estimator $\hat \psi \defeq {{\psi(W_{1:T}, \hat \eta)}}$.
            
        \STATE Return $\hat \psi$.
    \end{algorithmic}
    \end{algorithm}

    We will need several assumptions about the data generating process and nuisance function estimation.
    These assumptions are analogous to those necessary for the results in \citet{chernozhukov_doubledebiased_2018}.
    The first assumption, \cref{assm:regularity}, requires that $\psi$ is a smooth function of $\eta$, is Neyman orthogonal and satisfies regularity conditions. 
    We state it formally next.
    \begin{assumption}[Smoothness and Neyman orthogonality conditions]\label{assm:regularity}
        For all $T \geq 1$, the following conditions hold for all $P \in \cP$:
            \begin{enumerate}[(a)]
                \item  \label{item:twicegd} The map $\eta \mapsto \expt{\psi(W_{1:T}; \eta)}$ is twice continuously Gateaux-differentiable on $S_T$ around $\eta^*$.
                \item \label{item:neymanorth} The estimator $\psi$ is Neyman orthogonal with respect to the nuisance realization set $S_T$ around $\eta^*$.
            \end{enumerate}
        \end{assumption}
    \noindent \Cref{assm:regularity}\cref{item:twicegd} requires that second Gateaux derivatives of $\psi$ with respect to $\eta \in S_t$ exist and are continuous.
    The Gateaux derivative is defined in \cref{def:gateaux} and, informally, is equivalent to the usual derivative with respect to $r \in (0,1)$ for $\psi(W_{1:T}; r \eta^* + (1-r) \eta) $ for $\eta \in S_T$.
    \Cref{assm:regularity}\cref{item:neymanorth} requires Neyman orthogonality, which is defined in \cref{def:neymanorth} and, informally, is the requirement that the Gateaux derivative above, evaluated at $r=0$, is zero. {In applications, using our theorem in this section will require carefully designing estimators so that these conditions hold.}

    The second assumption, \cref{assm:rates}, requires that the nuisance estimators obey regularity conditions and converge at appropriate rates, for all $\eta \in S_T$.
        \begin{assumption}[Regularity and  convergence rate conditions] \label{assm:rates}
            There exists constants $\gamma, \delta, C > 0$ and $T_0 \geq 1$ such that for all $T > T_0$, the nuisance function estimator $\hat\eta (\Waux)$ belongs to a realization set $S_T$  with probability at least $1-\gamma$ where $S_T$ contains $\eta^*$ and satisfies the following conditions for all $P \in \cP$: 
                \begin{align}
                    &\sup_{\eta \in S_T} \norm{ \varphi(W_t; \eta)}_{L^{4+\delta}(P)} < C, \; \forall t \in T  \label{assm:2.1} \\
                    &\sup_{\eta \in S_T} T^{-1} \sum_{t=1}^T \norm{ \varphi(W_t; \eta) - \varphi(W_t; \eta^*)}_{L^2(P)} = o(1), \label{assm:2.2} \\
                    &\sup_{r \in(0,1), \eta \in S_T}\abs{ \partial^{(2)}_r \expt{ \psi(W_{1:T}; \eta^* + r(\eta - \eta^*))}} = o({T^{-1/2}}) \label{assm:2.3} 
                \end{align}
        \end{assumption}
    \noindent \Cref{assm:2.1} requires that the $L^q$ norm of $\varphi(W_t, \eta)$ must be bounded for all $t \in [T]$ for $q > 4$.
    \Cref{assm:2.2} requires the average over $t$ of the $L^2$ norm of $\varphi(W_t; \eta)-\varphi(W_t, \eta^*)$ must converge to zero as $T \to \infty$.
    Finally, \Cref{assm:2.3} requires the second Gateaux derivative of $\psi$ with respect to $\eta$ must go to zero at faster-than-$\sqrt{T}$ rates.
    These are analogous to standard regularity and convergence rate conditions for DML theorems \citep{chernozhukov_doubledebiased_2018}.

%

We are now ready to state our theorem, which says that the assumptions above are sufficient for efficient inference and consistent variance estimation under shared-state interference.
%

\begin{theorem}\label{thm:main}
    Under \Cref{assm:markovchain,assm:regularity,assm:rates},  with probability no less than $1 - \gamma$, \Cref{alg:dmlssv} returns an estimator such that \begin{align}
        \sqrt{T}\sigma^{-1}(\psi(W_{1:T}; \hat \eta) - \psi^*) \convindist N(0,1) \label{eq:dmlssv}
    \end{align}
    where we define $\sigma^2 = \lim_{T \to \infty} T \cdot \Var_P(\psi(W_{1:T}, \hat \eta))$.
    Moreover, when the variance of the estimator $\sigma^2$  is replaced with the variance estimator $\hat \sigma^2$ given below, \cref{eq:dmlssv} still holds:
    \begin{align}
        \hat \sigma^2 \defeq \begin{cases}
            \begin{aligned}
            \frac{1}{T_2(T_1 - 1)} \sum_{t=0}^{T_1-1} \bigg(\sum_{s=tT_2 + 1}^{(t+1) T_2} \varphi(W_s; \hat \eta) - \psi(W_{1:T}; \hat \eta)\bigg)^2 && \quad\quad\quad \text{under \cref{assm:markovchain}\cref{item:geoerg}}  
            \end{aligned} &  \\
            \begin{aligned}
               \frac{1}{T}\sum_{t=1}^T &(\varphi(W_{t}; \hat \eta) - \psi(W_{1:T}; \hat \eta))^2 + 2\sum_{i=1}^{m} (\varphi(W_{t}; \hat \eta) - \psi(W_{1:T}; \hat \eta)) (\varphi(W_{t-i}; \hat \eta) - \psi(W_{1:T}; \hat \eta))  
            \end{aligned} \\
            \quad\quad\quad\quad\quad\quad\quad\quad\quad\quad\quad\quad\quad\quad\quad\quad\quad\quad\quad\quad\quad\quad\quad\quad\quad \text{under \cref{assm:markovchain}\cref{item:mdep}}
        \end{cases}
        \label{eq:sigmasqhatdef}
    \end{align}
    where, for some $\theta > (1 + \delta/2)^{-1}$, we define $T_2 = \floor{T^{\theta}}$ and 
    $T_1 = \floor{T/T_2}$.
\end{theorem}

%
We briefly comment on the fact that our variance estimators under \cref{assm:markovchain}\cref{item:geoerg} and under \cref{assm:markovchain}\cref{item:mdep} are different.
Under \cref{assm:markovchain}\cref{item:geoerg}, our consistent variance estimator is not a plug-in estimator. 
In fact, plug-in variance estimators in Markov chains are not in general consistent \citep{jones_fixed-width_2006}. 
On the other hand, for \cref{assm:markovchain}\cref{item:mdep}, the plug-in variance estimator \textit{is} consistent.

The proof of our \cref{thm:main} follows the pattern in that of Theorem 3.1 of \citet{chernozhukov_doubledebiased_2018}.
We have simplified some of their setting for the sake of space and clarity, but our analysis in this paper could be extended to accommodate the additional generality in their theorems.

To prove the result, we must account for covariance between different terms $\varphi(W_t)$ in our analysis; \citet{chernozhukov_doubledebiased_2018} do not have to handle any covariances since each observation is independent.
To control covariance, we appeal to either the combination of geometric ergodicity and detailed balance or $m$-dependence.
In either of these cases, correlation between terms goes to zero sufficiently fast that these terms do not dominate.
After these covariances are handled, we may replace the iid central limit theorem used in \citet{chernozhukov_doubledebiased_2018} with an appropriate Markov chain central limit theorem to complete the result.
To prove consistency of the variance estimates, under \cref{assm:markovchain}\cref{item:geoerg} we apply a consistent variance estimation theorem from the Markov chain literature.
Under \cref{assm:markovchain}\cref{item:mdep}, we show that the plug-in variance estimator is consistent by showing the plug-in estimates of each variance and covariance term for $\varphi(W_t; \eta)$ are themselves consistent.

\paragraph{Applying the theorem.}
At a high level, instantiating \cref{thm:main} in specific settings is a matter of verifying \Cref{assm:markovchain,assm:regularity,assm:rates}. 
Once these are satisfied, \Cref{thm:main} tells us that the DML procedure in \Cref{alg:dmlssv} can be used to generate asymptotically normal treatment effect estimators using the provided variance estimators.

In the next two sections, we apply \cref{thm:main} for two structural models of independent interest. 
In each section, we examine a specific semiparametric structural model of outcomes given covariates, treatment and the shared state and a specific causal estimand. 
%
%
In each case, verifying \cref{assm:markovchain,assm:regularity,assm:rates} amounts to ensuring all interference is channeled through a shared state,
 making appropriate regularity assumptions and proving the following three lemmas:
first, that the estimator is Neyman orthogonal with respect to the nuisance estimators;
second, that the average over $t$ of the $L^2$ norm of $\varphi(W_t; \eta)-\varphi(W_t, \eta^*)$ must converge to zero as $T \to \infty$;
third, that the second Gateaux derivative of the estimator has order smaller than $\sqrt{T}$.
The regularity assumptions amount to support conditions requiring a lower bound on the probabilities of each treatment assignment and requirements that nuisance estimates and parameters are bounded in probability.
%

\ifproofsinbody
\noindent \textit{Proof \Cref{thm:main}.}
    Observe:
    \begin{align}
        \psi(W_{1:T}; \hat \eta) - \psi^*   
        &\leq  \psi(W_{1:T}; \eta^*) - \psi^* +  \abs{\psi(W_{1:T}; \hat \eta)) - \psi(W_{1:T}; \eta^*))} \nonumber \\
        &\leq \psi(W_{1:T}; \eta^*) - \psi^* \label{eq:oracle_diff}  \\
        &\quad + \big| {\psi(W_{1:T}; \hat \eta) - \psi(W_{1:T}; \eta^*) - \E_{P}[\psi(W_{1:T}; \hat \eta) - \psi(W_{1:T}; \eta^*) \; | \; \Waux]}   \big| \label{eq:empiricalproc}\\
        &\quad + \big| \E_{P}\sqparen{\psi(W_{1:T}; \hat \eta) - \psi(W_{1:T}; \eta^*) \; | \; \Waux} \big| \label{eq:plugin_vs_oracle} 
    \end{align}
    from applying the triangle inequality twice.
    Then applying \Cref{lem:empiricalproc} and \Cref{lem:neymanorth} implies
    \begin{align*}
        \sqrt{T} \sigma^{-1} \paren{\psi(W_{1:T}; \hat \eta) - \psi^*} = \sqrt{T} \sigma^{-1} (\psi(W_{1:T}; \eta^*) - \psi^*) +  o_{P}(1)
    \end{align*}
    Applying \Cref{lem:score}, we observe
    \begin{align*}
        \sqrt{T} \sigma^{-1} (\psi(W_{1:T}; \eta^*) - \psi^*) \convindist N(0, 1).
    \end{align*}
    which, with, e.g., \cite{van_der_vaart_asymptotic_2000} Theorem 2.7(iv), implies the first statement in the result.
    Lastly, \Cref{lem:var} proves
    $\hat \sigma^2_T \convinprob \sigma^2,$ and applying Slutsky's theorem completes the proof.
\qed

In the proof of \Cref{lem:score}, we will use the following two central limit theorems. The first is for geometrically ergodic Markov chains on general state spaces, and the second is for $m$-dependent sequences.
\Cref{thm:clt} generalizes standard central limit theorems for Markov chains in finite state spaces, where the analogous requirements are that the chain is irreducible and aperiodic.
See \Cref{sec:background} for further description of geometrically ergodic Markov chains.
\Cref{thm:mdepclt} states a central limit theorem for $m$-dependent sequences.

\begin{theorem}[Theorem 2, \cite{chan_discussion_1994}] \label{thm:clt}
    Suppose that a sequence of random variables $\{ A_t \}_{t=1}^\infty$ is a geometrically ergodic Markov chain with stationary distribution $\pi$.
    Also, suppose that there exists a constant $\delta > 0$ so that a measurable function $f$ satisfies $\norm{f}_{L^{2 + \delta}(\pi)} < \infty$. 
    Then it holds
    \begin{align*}
        \frac{1}{\sqrt{T} \sigma } \paren{\sum_{t=1}^{\infty} f(A_t) - \E_{\pi}[f]} \convindist N(0, 1),
    \end{align*}
    where we define
    \begin{align*}
        \sigma^2 \defeq \Var_\pi(f(A_1)) + 2 \sum_{t=1}^\infty \Cov_\pi(f(A_1), f(A_t)).
    \end{align*}
\end{theorem}

\begin{theorem}[Theorem 1, \cite{hoeffding_central_1948}]\label{thm:mdepclt}
    For a sequence of $m$-dependent random variables $A_1, A_2, \cdots$ if there exists constant $C > 0$, satisfying, for all $t=1,2,\dots$, $\mathbb{E}_P{\abs{A_t}^3} \leq C$.
    Then, 
    \begin{align*}
        T^{-1/2}\sigma^{-1} \sum_{t=1}^T (A_t - \ex{A_t}) \convindist N(0,1),
    \end{align*}
    where we define 
    \begin{align*}
        \sigma^2 &\defeq \lim_{\ell \to \infty} \ell^{-1} \sum_{s=1}^\ell \paren{\Var_P(A_{t+s}) + 2\sum_{i=1}^m \Cov_P(A_{t+s}, A_{t+s-i})}.
    \end{align*}
\end{theorem}

\Cref{lem:score} says that the oracle estimator approaches the estimand at $\sqrt{T}$ rates and is asymptotically normal.
The proof of \cref{lem:score} applies \Cref{thm:clt}. To apply the CLT, we just need to verify that we can apply one of the two central limit theorems above: \cref{thm:clt} for \cref{assm:markovchain}\cref{item:geoerg} and \cref{thm:mdepclt} for \cref{item:mdep}.

\begin{lemma}[Corollary to \Cref{thm:clt}] \label{lem:score}
    For an estimator $\psi$ satisfying \Cref{assm:markovchain} and \Cref{assm:rates}, it holds
    \begin{align*}
        \sqrt{T} \sigma^{-1}(\psi(W_{1:T}; \eta^*) -\psi^*) \convindist N(0, 1).
    \end{align*}
\end{lemma}

\noindent \textit{Proof of \Cref{lem:score}.} 
    From \Cref{assm:markovchain}, under part \cref{item:geoerg}, we have that $\{ W_t \}$ is a geometrically ergodic Markov chain. 
    Also, $\varphi(W_t; \eta^*)$ is bounded in $L^{2+\delta}(P)$ by \cref{assm:rates}, \cref{assm:2.1}. 
    So $\psi(W_{1:T}; \eta^*) - \psi^*$ is bounded in probability by the triangle inequality: 
    \begin{align*}
        \norm{\psi(W_{1:T}; \eta^*) - \psi^*}_{L^{2+\delta}(P)} \leq T^{-1} \sum_{t=1}^T \norm{\varphi(W_{t}; \eta^*)}_{L^{2+\delta}(P)} + \abs{\psi^*} < \infty
    \end{align*}
    Thus, the result under \cref{assm:markovchain}\cref{item:geoerg} follows directly from the application of \Cref{thm:clt}.

    For \Cref{assm:markovchain}\cref{item:mdep}, since $\{ W_t \}_{t=1}^\infty$ is $m$-dependent, so is $\{ \psi(W_t; \hat \eta) \}_{t=1}^\infty$.
    By \cref{assm:rates}, \cref{assm:2.1}, we have that $\varphi(W_{t}; \eta^*)$ is bounded in $L^3$.
    Thus, we can apply \cref{thm:mdepclt}.
    
\qed

\vspace{1em}

\Cref{lem:empiricalproc} says that the deviation from its mean of the difference between the oracle estimator and our estimator approaches zero at faster-than-$\sqrt{T}$ rates.
Our proof pattern is similar to that of \citet{chernozhukov_doubledebiased_2018}: our strategy is to prove that the variance of the expression goes to zero at faster-than-$T$ rates, so that we can apply Chebyshev's inequality, which implies that the expression itself goes to zero at faster-than-$\sqrt{T}$ rates.
The core difficulty of proving the lemma, compared to the analogous result in \citet{chernozhukov_doubledebiased_2018}, is that we must account for the covariance between observations at different times.
\citet{chernozhukov_doubledebiased_2018} assumes independence between observations and thus does not have to account for such covariances.
We handle the covariances separately for \Cref{assm:markovchain}\cref{item:geoerg} and for \Cref{assm:markovchain}\cref{item:mdep}.
Under \Cref{assm:markovchain}\cref{item:geoerg}, we can invoke a theorem saying that geometrically ergodic Markov chains satisfying detailed balance have correlations between observations that decrease to zero at an exponential rate.
This implies that the sum of correlations is bounded by a constant, so they do not dominate.
Under \Cref{assm:markovchain}\cref{item:mdep}, there are only finitely many non-zero correlations, so their sum is also bounded by a constant.

\begin{lemma} \label{lem:empiricalproc} 
    Under \cref{assm:markovchain}, for an estimator $\psi$ and nuisance parameter estimators $\hat \eta$ satisfying \Cref{assm:regularity} and \Cref{assm:rates}, with probability $1 - \gamma$, it holds
    \begin{align*}
        \big| {\psi(W_{1:T}; \hat \eta) - \psi(W_{1:T}; \eta^*) - \E_{P}[\psi(W_{1:T}; \hat \eta) - \psi(W_{1:T}; \eta^*) \; | \; \Waux]}   \big| = o_{P}(T^{-1/2})
    \end{align*}
\end{lemma}

\noindent \textit{Proof of \Cref{lem:empiricalproc}.} 
    First, we will introduce new notation. For all $t \in [T]$, let
    \newcommand{\phihat}{\hat\varphi_t}
    \newcommand{\tphihat}[1]{\hat\varphi_{#1}}
    \newcommand{\phistar}{{\varphi_t^*}}
    \newcommand{\tphistar}[1]{\varphi_{#1}^*}
    \begin{align*}
        \phihat \defeq \varphi(W_{t}; \hat \eta), \\
        \phistar \defeq \varphi(W_{t}; \eta^*).
    \end{align*}
    Let $\cE_T$ be the event that $\hat \eta \in S_T$ (which happens with probability $1 - \gamma$ by \cref{assm:rates}).
    Also, $\hat\eta$ is assumed to be a deterministic function of $\Waux$.
    On the event $\cE_T$,
    \begin{align}
        &\E_{P} \sqparen{(\psi(W_{1:T}; \hat \eta) - {\psi(W_{1:T}; \eta^*) - \E_{P}[\psi(W_{1:T}; \hat \eta) - \psi(W_{1:T}; \eta^*) \; | \; \Waux]})^2\; | \; \Waux} \nonumber \\
        &=\E_{P} \sqparen{\paren{T^{-1} \sum_{t=1}^T \phihat - {\phistar - \E_{P}[\phihat - \phistar \; | \; \Waux]}}^2\; | \; \Waux} \tag{Definition of $\psi$.} \\
        &={T^{-2} \sum_{t=1}^T \E_{P} [\paren{\phihat - {\phistar - \E_{P}[\phihat - \phistar \; | \; \Waux]}}^2} \; | \; \Waux] \nonumber \\
        &\quad\quad\quad\quad + 2 \sum_{\ell=1}^{t-1} \E_{P} \big[ \paren{\phihat - {\phistar - \E_{P}[\phihat - \phistar \; | \; \Waux]}} \\
        &\quad\quad\quad\quad\quad\quad\quad\quad\quad\cdot \paren{\tphihat{t-\ell} - {\tphistar{t-\ell} - \E_{P}[\tphihat{t-\ell} - \tphistar{t-\ell} \; | \; \Waux]}} \; \big| \; \Waux\big] \tag{Rearranging.}\\
        &\leq \sup_{\eta \in S_T} {T^{-2} \sum_{t=1}^T \E_{P} [\paren{\varphi(W_{t};  \eta) - {\phistar - \E_{P}[\varphi(W_{t};  \eta) - \phistar \; | \; \Waux]}}^2} \; | \; \Waux] \nonumber \\
        &\quad\quad\quad\quad\quad\quad + 2 \sum_{\ell=1}^{t-1} \E_{P} \big[ \paren{\varphi(W_{t};  \eta) - {\phistar - \E_{P}[\varphi(W_{t};  \eta) - \phistar \; | \; \Waux]}} \nonumber \\
        &\quad\quad\quad\quad\quad\quad\quad\quad\quad\quad\quad \cdot \paren{\varphi(W_{t-\ell};  \eta) - {\tphistar{t-\ell} - \E_{P}[\varphi(W_{t-\ell};  \eta) - \tphistar{t-\ell} \; | \; \Waux]}} \; \big| \; \Waux\big] \tag{$\hat \eta \in S_T$ on $\cE_T$} \\
        &\begin{aligned}
        &= \sup_{\eta \in S_T} {T^{-2} \sum_{t=1}^T \E_{P} [\paren{\varphi(W_{t};  \eta) - {\phistar - \E_{P}[\varphi(W_{t};  \eta) - \phistar ]}}^2} ]  \\
        &\quad\quad\quad\quad\quad\quad + 2 \sum_{\ell=1}^{t-1} \E_{P} \big[ \paren{\varphi(W_{t};  \eta) - {\phistar - \E_{P}[\varphi(W_{t};  \eta) - \phistar ]}}  \\
        &\quad\quad\quad\quad\quad\quad\quad\quad\quad\quad\quad\cdot \paren{\varphi(W_{t-\ell};  \eta) - {	\tphistar{t-\ell} - \E_{P}[\varphi(W_{t-\ell};  \eta) - 	\tphistar{t-\ell}]}} \big]  
        \end{aligned}\label{eq:covterm}
    \end{align}
    where the last equality comes from the {independence of $\Waux$ from $W_{1:T}$.}
    Now, for all $t \in \N$, let $\cF_{t}^\infty$ be the $\sigma$-algebra generated by $ W_{t:\infty}$ and let $\cF_1^t$ be the $\sigma$-algebra generated by $W_{1:t}$.
    Recall the definition of $\rho$-mixing (see, e.g., \cite{bradley_chapter_2007}): 
    \begin{align*}
        \rho(t) \defeq \sup_{f \in L^2(\cF_{i}^{\infty}), g \in L^2(\cF_1^{i-t})} \abs{\Corr(f,g)},
    \end{align*}
    where the correlation, covariance and variance for functions has the usual definition:
    \begin{align*}
    &\Corr(f,g) = \Cov(f, g)/(\Var(f)\Var(g))^{1/2}, \\
        &\Cov(f,g) = \expt{(f(W_{i:\infty}) - \expt{f(W_{i:\infty})})(g(W_{1:(i-t)})) - \expt{g(W_{1:(i-t)})})},
    \end{align*}
    and $\Var(f) = \Cov(f,f)$.
    Next, we state a lemma that will allow us to bound the covariance between terms in the above sum under \cref{assm:markovchain}\cref{item:geoerg}.

    \begin{lemma}[Theorem 2, \cite{jones_markov_2005}] \label{lem:rhomixing}
        If a Markov chain $A_{1:\infty}$ is geometrically ergodic and satisfies detailed balance, then it is $\rho$-mixing with $\rho(T) \leq O(e^{-c T})$ for some $c > 0$.
    \end{lemma}

    On the other hand, under \cref{assm:markovchain}\cref{item:mdep}, $W_{1:\infty}$ is $\rho$ mixing with $\rho(\ell) = 0$ for all $\ell > m$.
    Note that these facts imply $W_{1:\infty}$ is $\rho$-mixing with $\rho(t) = O(e^{-c T})$ for some $c > 0$.
    Also, for all $t \in [T]$, $\eta \in S_T$ and $\ell < t$, it holds ${\varphi(W_{t};  \eta) - {\phistar - \E_{P}[\varphi(W_{t};  \eta) - \phistar ]}} \in L^2(\cF_{t}^\infty)$ and ${\varphi(W_{t-\ell};  \eta) - {	\tphistar{t-\ell} - \E_{P}[\varphi(W_{t-\ell};  \eta) - 	\tphistar{t-\ell}]}} \in L^2(\cF_{1}^{t-\ell})$ (where expectations are taken with respect to $P$). 
    To see this, note that for all $t \in [T]$ and $\eta \in S_T$, it holds $\norm{{\varphi(W_{t};  \eta)}}_{L^2(P)} < \infty$ by \cref{assm:rates}, \cref{assm:2.1}.
    Also, since the expressions just depend on $W_t$ and $W_{t-\ell}$, respectively, this implies $\norm{{\varphi(W_{t};  \eta)}}_{L^2(P)} = \norm{{\varphi(W_{t};  \eta)}}_{L^2(\cF_{t}^\infty)}$ and similarly $\norm{{\varphi(W_{t-\ell};  \eta)}}_{L^2(P)} = \norm{{\varphi(W_{t-\ell};  \eta)}}_{L^2(\cF_1^{t-\ell})}$.
    Also, it is assumed $\eta^* \in S_T$, so by the same reasoning $\norm{{\varphi^*_t}}_{L^2(\cF_{t}^\infty)} < \infty$ and $\norm{{\varphi^*_{t-\ell}}}_{L^2(\cF_1^{t-\ell})} < \infty$.

    Thus, 
    \begin{align*}
        \Corr \paren{{\varphi(W_{t};  \eta) - {\phistar - \E_{P}[\varphi(W_{t};  \eta) - \phistar ]}}, {\varphi(W_{t-\ell};  \eta) - {	\tphistar{t-\ell} - \E_{P}[\varphi(W_{t-\ell};  \eta) - 	\tphistar{t-\ell}]}}} \leq \rho(\ell).
    \end{align*}
    This implies we can rewrite \cref{eq:covterm} as 
    \begin{align*}
        &\sup_{\eta \in S_T} {T^{-2} \sum_{t=1}^T \E_{P} [\paren{\varphi(W_{t};  \eta) - {\phistar - \E_{P}[\varphi(W_{t};  \eta) - \phistar ]}}^2} ] \nonumber \\
        &\quad\quad\quad\quad\quad\quad + 2 \sum_{\ell=1}^{t-1} \E_{P} \big[ \paren{\varphi(W_{t};  \eta) - {\phistar - \E_{P}[\varphi(W_{t};  \eta) - \phistar ]}} \\
        &\quad\quad\quad\quad\quad\quad\quad\quad\quad\quad\cdot \paren{\varphi(W_{t-\ell};  \eta) - {	\tphistar{t-\ell} - \E_{P}[\varphi(W_{t-\ell};  \eta) - 	\tphistar{t-\ell}]}} \big]  
        \\
        &\leq  \sup_{\eta \in S_T} T^{-2} \sum_{t=1}^T \norm{{\varphi(W_{t};  \eta) - {\phistar - \E_{P}[\varphi(W_{t};  \eta) - \phistar ]}}}^2_{L^2(P)}  \\
        &\quad\quad\quad\quad\quad\quad + 2 \sum_{\ell=1}^{t-1} \rho(\ell) \norm{{\varphi(W_{t};  \eta) - {\phistar - \E_{P}[\varphi(W_{t};  \eta) - \phistar ]}}}_{L^2(P)} \\
        &\quad\quad\quad\quad\quad\quad\quad\quad\quad \cdot \norm{{\varphi(W_{t-\ell};  \eta) - {\phistar - \E_{P}[\varphi(W_{t-\ell};  \eta) - \varphi^*_{t-\ell}]}}}_{L^2(P)} \\
        &\leq  \sup_{\eta \in S_T} T^{-2} \sum_{t=1}^T \norm{{\varphi(W_{t};  \eta) - {\phistar }}}^2_{L^2(P)} \\
        &\quad\quad\quad\quad\quad\quad + 2 \sum_{\ell=1}^{t-1} \rho(\ell) \norm{{\varphi(W_{t};  \eta) - {\phistar }}}_{L^2(P)} \norm{{\varphi(W_{t-\ell};  \eta) - {\varphi_{t-\ell}^*}}}_{L^2(P)} 
    \end{align*}
    Now, notice $$T^{-1} \sum_{t=1}^T \norm{{\varphi(W_{t};  \eta) - {\phistar }}}^2_{L^2(P)} = o(1)$$ by \cref{assm:rates}, \Cref{assm:2.2}, so $$T^{-2} \sum_{t=1}^T \norm{{\varphi(W_{t};  \eta) - {\phistar }}}^2_{L^2(P)} = o(T^{-1}).$$
    Next, since $\rho(\ell) = O(e^{-c \ell})$ for some $c > 0$ (by \cref{lem:rhomixing}) and $$T^{-1} \sum_{t=1}^T \norm{{\varphi(W_{t};  \eta) - {\phistar }}}_{L^2(P)}= o(1)$$ for all $t \in [T]$, it holds
    \begin{align*}
        &T^{-2} \sum_{t=1}^T \sum_{\ell = 1}^{t-1} \rho(\ell) \norm{{\varphi(W_{t};  \eta) - {\phistar }}}_{L^2(P)} \norm{{\varphi(W_{t-\ell};  \eta) - {\tphistar{t-\ell}}}}_{L^2(P)} \\
        &={T^{-2} \sum_{\ell = 1}^T \rho(\ell)\sum_{t=\ell + 1}^T   \norm{{\varphi(W_{t-\ell};  \eta) - {\tphistar{t-\ell}}}}_{L^2(P)} \norm{{\varphi(W_{t};  \eta) - {\phistar }}}_{L^2(P)}} \tag{Exchanging sums.} \\
        &=T^{-2} \sum_{\ell = 1}^T \rho(\ell)\paren{\sum_{t=\ell + 1}^T   \norm{{\varphi(W_{t-\ell};  \eta) - {\tphistar{t-\ell}}}}_{L^2(P)}}^{1/2} \paren{\sum_{t=\ell + 1}^T \norm{{\varphi(W_{t};  \eta) - {\phistar }}}_{L^2(P)}}^{1/2} \tag{Cauchy-Schwarz inequality.} \\
        &=T^{-1} \sum_{\ell = 1}^T \rho(\ell)\paren{T^{-1} \sum_{t=\ell + 1}^T   \norm{{\varphi(W_{t-\ell};  \eta) - {\tphistar{t-\ell}}}}_{L^2(P)}}^{1/2} \paren{T^{-1} \sum_{t=\ell + 1}^T \norm{{\varphi(W_{t};  \eta) - {\phistar }}}_{L^2(P)}}^{1/2} \tag{Distributing $T^{-1}$.} \\
        &=T^{-1} \sum_{\ell = 1}^T \rho(\ell)o(1) \tag{${T^{-1} \sum_{t=\ell + 1}^T \norm{{\varphi(W_{t};  \eta) - {\phistar }}}_{L^2(P)}} = o(1)$.} \\
        &=T^{-1} \sum_{\ell = 1}^T O(e^{-c \ell}) o(1) \tag{$\rho(\ell) = O(e^{-c\ell})$ for some $c > 0$.} \\
        &\leq o(T^{-1}). 
    \end{align*}
    where the last line comes from the fact that {$\sum_{t=1}^T e^{-\theta t} \leq 1/(1- e^{-\theta})$}.
    Thus, applying Chebyshev's inequality, we have
    \begin{align*}
        {{\paren{\psi(W_{1:T}; \hat \eta) - {\psi(W_{1:T}; \eta^*) - \E_{P}[\psi(W_{1:T}; \hat \eta) - \psi(W_{1:T}; \eta^*) \; | \; \Waux]}}} \; | \; \Waux} = o_{P}(T^{-1/2}).
    \end{align*}
    Finally, the fact that conditional convergence implies unconditional convergence (Lemma 6.1 of \cite{chernozhukov_doubledebiased_2018}) implies the desired result. \qed
    
\vspace{1em}

\Cref{lem:neymanorth} says that the absolute value of the expectation of the difference between the estimator and oracle goes to zero at faster-than-$\sqrt{T}$ rates.
The proof is nearly identical to the analogous argument in \citet{chernozhukov_doubledebiased_2018}: we apply Taylor's theorem by differentiating on the path between $\hat \eta$ and $\eta^*$ to show that small deviations of nuisances $\eta$ around $\eta^*$ do not dramatically affect the deviation of the estimator from the oracle estimator.
The first-order term is zero by the assumption of Neyman orthogonality.
The second-order term is small by assumption (\Cref{assm:2.3}).

\begin{lemma} \label{lem:neymanorth}
    For an estimator $\psi$ and nuisance parameter estimators $\hat \eta$ satisfying \Cref{assm:regularity} and \Cref{assm:rates}, it holds 
    \begin{align*}
         \big| \E_{P}\sqparen{\psi(W_{1:T}; \hat \eta) - \psi(W_{1:T}; \eta^*) \; | \; \Waux} \big| = o_{P}(T^{-1/2}).
    \end{align*}
\end{lemma}

\noindent \textit{Proof of \Cref{lem:neymanorth}.}  
    \noindent We use Neyman orthogonality. Define
    \begin{align*}
    f(r) \defeq {\E_{P}\sqparen{\psi(W_{1:T}; \eta^* + r(\hat \eta - \eta^*)) \; | \; \Waux} - \E_{P}\sqparen{\psi(W_{1:T}; \eta^*)}}, && r \in [0, 1]
    \end{align*}
    Then, notice
    \begin{align*}
        f(1) = {\E_{P}\sqparen{\psi(W_{1:T}; \hat \eta) \; | \; \Waux} - \E_{P}\sqparen{\psi(W_{1:T}; \eta^*)}}
    \end{align*}
    is the quantity we want to bound in probability.
    By Taylor's Theorem
    \begin{align*}
        f(1) &= f(0) + \frac{\partial}{\partial r} f(0) + \frac{\partial^2}{\partial r^2}f(\tilde r)/2
    \end{align*}
    for some $\tilde r \in (0,1)$.
    Notice that $f(0) = 0$ since $\psi(W_{1:T}; \eta^*)$ is independent of $\Waux$.
    Also, notice that $\partial f(0) / \partial r = 0$ by \Cref{assm:regularity}\ref{item:neymanorth}.
    Finally, on the event $\cE_T$,
    \begin{align*}
        \abs{\frac{\partial^2}{\partial r^2} f(\tilde r)} \leq \sup_{r \in (0,1)} \abs{\frac{\partial^2}{\partial r^2}f(r)} = o_{P}(T^{-1/2})
    \end{align*}
    where the convergence comes from \Cref{assm:2.3}.
\qed

\vspace{1em}

\Cref{lem:var} states the consistency of our variance estimator $\hat \sigma^2$.
We prove consistency for the estimators under \Cref{assm:markovchain}\cref{item:geoerg} and \Cref{assm:markovchain}\cref{item:mdep} separately.
For \Cref{assm:markovchain}\cref{item:geoerg}, we can directly apply a consistent variance estimation result from prior work.
For \Cref{assm:markovchain}\cref{item:mdep}, we prove consistency of the plugin variance estimator directly. We handle the variance terms and covariance terms in $\psi$ separately. For the variance terms, we follow the argument in \citet{chernozhukov_doubledebiased_2018}. For the covariance terms, we follow a similar pattern to the argument for the variance terms: show that the average of the covariances converge in probability to covariance estimates computed using oracle nuisances, and then showing that the covariance estimates computed using oracle nuisances converge to their expectation.

\begin{lemma} \label{lem:var}
    For $\hat \sigma_T^2$ as defined in \Cref{eq:sigmasqhatdef} and $\sigma^2 = \lim_{t \to \infty} \Var_P(\psi(W_{1:T}, \hat \eta)$ and under \Cref{assm:markovchain,assm:regularity,assm:rates}, it holds
        $\hat \sigma_T^2 \convinprob \sigma^2.$
\end{lemma}

\noindent \textit{Proof.}
We split the analysis into two cases. The first case handles when \cref{assm:markovchain}\cref{item:geoerg} holds and the second handles when \cref{assm:markovchain}\cref{item:mdep} holds.

\paragraph{Case 1: Under \cref{assm:markovchain}\cref{item:geoerg}.}
We will apply the following result adapted from \cite{jones_fixed-width_2006}.
For simplicity, we will use $T_1, T_2$ as defined in the statement of the result.
\begin{lemma}[Proposition 3, \cite{jones_fixed-width_2006}] \label{lem:geoerg_clt}
    Suppose a Markov chain $\{ A_t \}_{t=1}^\infty$ is geometrically ergodic with stationary distribution $\pi$. Also suppose that there exists a constant $\delta$ such that a measurable function $f$ satisfies $\norm{f}_{L^{2+\delta}(\pi)} < C$. 
    Define 
    \begin{align*}
        \sigma^2 \defeq \Var_{\pi}(f(A_1)) + 2 \sum_{t=1}^\infty \Cov_{\pi}(f(A_1), f(A_t)). 
    \end{align*}
    and
    \begin{align*}
        \hat \sigma^2 \defeq \frac{1}{T_2(T_1 - 1)} \sum_{t=0}^{T_1-1} \bigg(\sum_{s=tT_2 + 1}^{(t+1) T_2} f(A_t) - T^{-1} \sum_{i=1}^T f(A_t)\bigg)^2 .
    \end{align*}
    Then it holds $\hat \sigma^2 \to \sigma^2$.
\end{lemma}
Notice that by \cref{assm:rates}, \cref{assm:2.1} and the fact that $K_\infty \in \cP$, we have assumed $\norm{\varphi(W_t; \eta)}_{L^{4+\delta}(K^\infty)} < C$.
Thus, we can apply \Cref{{lem:geoerg_clt}} and the result under \cref{assm:markovchain}\cref{item:geoerg} follows.

\paragraph{Case 2: Under \cref{assm:markovchain}\cref{item:mdep}}
Define
    \begin{align*}
        \sigma^2_T \defeq \sum_{t=1}^{T-m} \Var(\varphi(W_{m+t}; \eta^*)) + 2\sum_{i=1}^m \Cov(\varphi(W_{m+t}; \eta^*), \varphi(W_{m+t-i}; \eta^*))
    \end{align*}
    We will prove
    \begin{align*}
        \hat \sigma^2_T - \sigma_T^2 \convinprob 0
    \end{align*}                                                     which will imply the result, since $\sigma_T^2 \to \sigma^2$ trivially. 
Using the notation from \Cref{lem:empiricalproc}, notice
    \begin{align}
        \hat \sigma^2_T - \sigma_T^2 &= T^{-1} \sum_{t=1}^T \paren{\tphihat{t} - \psi(W_{1:T}; \hat \eta) }^2 - \ex{(\tphistar{t} - \psi^*)^2} \label{eq:varterm} \\
        &\quad + 2T^{-1} \sum_{t=1}^T \sum_{i=1}^{\min \{ t - 1, m \}} \paren{\tphihat{t} - \psi(W_{1:T}; \hat \eta) } \paren{\tphihat{t-i} - \psi(W_{1:T}; \hat \eta) } - \ex{(\tphistar{t} - \psi^*)(\tphistar{t-i} - \psi^*)} \label{eq:covterm2}
    \end{align}
    by definition.
    We will bound \cref{eq:varterm} and \cref{eq:covterm2} separately.
    First, for \cref{eq:varterm},
    \begin{align}
        &T^{-1} \sum_{t=1}^T \paren{\phihat - \psi(W_{1:T}; \hat \eta) }^2 - \ex{(\phistar - \psi^*)^2} \nonumber \\
        &= T^{-1} \sum_{t=1}^T\paren{\phihat - \psi(W_{1:T}; \hat \eta) }^2 - \paren{\phistar - \psi^* }^2 \label{eq:hatminusstar} \\
        &\quad\quad\quad\quad + \paren{\phistar - \psi^* }^2 - \ex{\paren{\phistar - \psi^* }^2} \label{eq:starminusex}
    \end{align}
    by adding and subtracting terms.
    Next, to bound \cref{eq:hatminusstar} in probability, notice
    \begin{align}
        &{\paren{T^{-1} \sum_{t=1}^T\paren{\phihat - \psi(W_{1:T}; \hat \eta) }^2 - \paren{\phistar - \psi^* }^2}^2} \nonumber \\
        &= T^{-2} {\paren{\sum_{t=1}^T\paren{\phihat - \psi(W_{1:T}; \hat \eta)  - \phistar + \psi^* }\paren{\phihat - \psi(W_{1:T}; \hat \eta)  + \phistar - \psi^* }}^2} \nonumber \\
        &\leq T^{-1}   {\sum_{t=1}^T\paren{\phihat - \psi(W_{1:T}; \hat \eta)  - \phistar + \psi^*}^2} \label{eq:diffterm}\\
        &\quad\cdot T^{-1}{\sum_{t=1}^T \paren{\phihat - \psi(W_{1:T}; \hat \eta)  + \phistar - \psi^* }^2} \label{eq:sumterm}
    \end{align}
    The first (in)equality comes from rearranging; the second comes from the Cauchy-Schwarz inequality.
    Now, we can bound \cref{eq:diffterm} as
    \begin{align}
        \nonumber &T^{-1} {\sum_{t=1}^T\paren{\phihat - \psi(W_{1:T}; \hat \eta)  - \phistar + \psi^* }^2} \\
        \nonumber &\leq 2T^{-1} \paren{\sum_{t=1}^T \paren{\phihat - \phistar}^2} + 2\paren{\psi(W_{1:T}; \hat \eta) - \psi^*}^2 \\
        \label{eq:diffterm1} &\leq 2T^{-1} \paren{\sum_{t=1}^T \paren{\phihat - \phistar}^2} \\
        \label{eq:diffterm2}& \quad + 4\paren{\psi(W_{1:T}; \hat \eta)  - \expt{\psi(W_{1:T}; \hat \eta) \; | \; \Waux}}^2 \\
        \label{eq:diffterm3}& \quad + 4\expt{\psi(W_{1:T}; \hat \eta) - \psi^* \; | \; \Waux}^2 
    \end{align}
    where the first inequality comes from the fact that $(a+b)^2 \leq 2(a^2 + b^2)$ for all $a,b$; the second comes from adding and subtracting the conditional expectation and then again applying $(a+b)^2 \leq 2(a^2 + b^2)$. 
    \Cref{eq:diffterm1} is $o_{P}(1)$ by Markov's inequality and \cref{assm:2.2} since
    \begin{align*}
        T^{-1} \paren{\sum_{t=1}^T \expt{\paren{\phihat - \phistar}^2}} &= T^{-1} \sum_{t=1}^T\norm{\phihat - \phistar}^2_{L^2(P)}\\
        &=T^{-1} \sum_{t=1}^T o(1).
    \end{align*}
    \Cref{eq:diffterm2} is $o_{P}(1)$ by the following sequence.
    \begin{align*}
        &\paren{\psi(W_{1:T}; \hat \eta)  - \expt{\psi(W_{1:T}; \hat \eta) \; | \; \Waux}}^2 \\
        &= T^{-2} \paren{\sum_{t=1}^T \phihat - \expt{\phihat}}^2 \tag{Definition of $\psi$.} \\
        &= T^{-2} \sum_{t=1}^T \paren{\phihat - \expt{\phihat}}^2 + 2\sum_{i=1}^m (\phihat - \expt{\phihat})(\tphihat{t-i} - \expt{\tphihat{t-i}}) \tag{Rearranging.} \\
        &\leq T^{-2} \sum_{t=1}^T \paren{\phihat - \expt{\phihat}}^2 + 2\sqrt{m} (\phihat - \expt{\phihat})\paren{\sum_{i=1}^m(\tphihat{t-i} - \expt{\tphihat{t-i}})^2}^{1/2} \tag{Cauchy-Schwarz inequality.} \\
        &= T^{-2} \sum_{t=1}^T O_{P}(1) + 2\sqrt{m} O_{P}(1) \paren{\sum_{i=1}^mO_{P}(1)}^{1/2}
    \end{align*}
    The last line comes by Markov's inequality and the fact that $\expt{\paren{\phihat - \expt{\phihat}}^2} = O(1)$ for all $t$ by \cref{assm:2.1} and the fact that $m$ is a constant independent of $T$.
    \Cref{eq:diffterm3} is $o(1)$ by the following sequence.
    \begin{align*}
        \expt{\psi(W_{1:T}; \hat \eta) - \psi^* \; | \; \Waux}^2 &= \expt{\psi(W_{1:T}; \hat \eta) - \psi(W_{1:T}; \eta^*) \; | \; \Waux}^2 \\
        &= T^{-2} \paren{\sum_{t=1}^T \expt{\phihat - \phistar} }^2 \\
        &=T^{-2} \paren{\sum_{t=1}^T o(1) }^2\\
        &= o(1)
    \end{align*}
    The first line comes by the fact that $\expt{\psi(W_{1:T}; \eta^*) \; | \; \Waux} = \psi^*$, the second is by definition of $\psi$, and the last is \cref{assm:2.2}.
    Thus, \cref{eq:diffterm} is $o_{P}(1)$.
    We can bound \cref{eq:sumterm} in probability as
    \begin{align}
        \nonumber &T^{-1}{\sum_{t=1}^T \paren{\phihat - \psi(W_{1:T}; \hat \eta)  + \phistar - \psi^* }^2} \\
        \nonumber &\leq 2 T^{-1}\paren{\sum_{t=1}^T \paren{\phihat + \phistar}^2} + 2 (\psi(W_{1:T}; \hat\eta) + \psi^*)^2\\
        \nonumber &\leq 4 T^{-1}\paren{\sum_{t=1}^T {{\phihat}^2 + {\phistar}^2}} + 4T^{-2} \paren{\sum_{t=1}^T {\phihat}}^2 + 4{\psi^*}^2\\
        \nonumber &\leq \sup_{\eta \in S_T} 4 T^{-1}\paren{\sum_{t=1}^T {\varphi(W_{t}; \eta)^2 + {\phistar}^2}} + 4T^{-2} \paren{\sum_{t=1}^T {\varphi(W_{t}; \eta)}}^2 + 4{\psi^*}^2\\
        &= O_{P}(1) \label{eq:sumterm_o}
    \end{align}
    where in each inequality, we apply the fact that $(a+b)^2 \leq 2a^2 + 2b^2$ and in the last line we apply Markov's inequality with \cref{assm:2.1}: Namely, from  \cref{assm:2.1} there exists a constant $C$ such that, for all $s, t \in [T]$,
    \begin{align*}
        \sup_{\eta \in S_T} \ex{\varphi(W_{t}; \eta)^2} = \norm{\varphi(W_{t}; \eta)}_{L^2(P)}^2 \leq C,& \\
        \ex{{\phistar}^2} = \norm{\phistar}_{L^2(P)}^2 \leq C,& \text{ and}\\
        \sup_{\eta \in S_T} \ex{\varphi(W_{t}; \eta) \varphi(W_{s}; \eta)} \leq \norm{\varphi(W_{t}; \eta)}_{L^2(P)}\norm{\varphi(W_{s}; \eta)}_{L^2(P)} \leq C.&  
    \end{align*}
    (The last line applies the Cauchy-Schwarz inequality.)
    The fact that \cref{eq:diffterm,eq:sumterm} are $o_{P}(1)$ and $O_{P}(1)$ respectively imply that the expression in \cref{eq:hatminusstar} is $o_{P}(1)$.
    Next, we bound \Cref{eq:starminusex} in probability. Notice
    \begin{align*}
        &\ex{\paren{T^{-1}\sum_{t=1}^T \paren{\phistar - \psi^* }^2 - \ex{\paren{\phistar - \psi^* }^2}}^2} \\
        &= T^{-2}\sum_{t=1}^T \ex{\paren{\paren{\phistar - \psi^*}^2 - \ex{\paren{\phistar - \psi^* }^2}}^2} \\
        &\quad\quad\quad\quad + \sum_{i=1}^m \ex{ \paren{\paren{\phistar - \psi^*}^2 - \ex{\paren{\phistar - \psi^* }^2}} \paren{{\paren{\tphistar{t-i} - \psi^*}^2 - \ex{\paren{\tphistar{t-i} - \psi^* }^2}}}} \\ 
        &\leq T^{-2}\sum_{t=1}^T \norm{{\phistar - \psi^*}}_{L^4(P)}^4 \\
        &\quad\quad\quad\quad + \sqrt{m} \norm{\phistar - \psi^*}_{L^2(P)}^2 \paren{\sum_{i=1}^m\norm{\tphistar{t-i} - \psi^*}_{L^2(P)}^2}^{1/2} \\
        &\leq T^{-2}\sum_{t=1}^T \norm{{\phistar - \psi^*}}_{L^4(P)}^4 \\
        &\quad\quad\quad\quad + \sqrt{m} \norm{\phistar - \psi^*}_{L^2(P)}^2 \paren{\sum_{i=1}^m\norm{\tphistar{t-i} - \psi^*}_{L^2(P)}^2}^{1/2} \\
        &= o_{P}(1)
    \end{align*}
    where in the last line we apply the triangle inequality $\norm{{\phistar - \psi^*}}_{L^4(P)} \leq \norm{{\phistar}}_{L^4(P)} + \abs{\psi^*}$ and \cref{assm:2.1}.
    Thus, the fact that \cref{eq:hatminusstar} is $o_{P}(1)$ and that \cref{eq:starminusex} is $o_{P}(1)$ imply that \cref{eq:varterm} is $o_{P}(1)$.
    To bound \cref{eq:covterm2},
    \begin{align}
        \nonumber&T^{-1} \sum_{t=1}^T \sum_{i=1}^m \paren{\phihat - \psi(W_{1:T}; \hat \eta) } \paren{\tphistar{t-i} - \psi(W_{1:T}; \hat \eta) } - \ex{(\phistar - \psi^*)(\tphistar{t-i} - \psi^*)} \\
        &=T^{-1} \sum_{t=1}^T \sum_{i=1}^m \paren{\phihat - \psi(W_{1:T}; \hat \eta) } \paren{\tphihat{t-i} - \psi(W_{1:T}; \hat \eta) } - \paren{\phistar - \psi^*} \paren{\tphistar{t-i} - \psi^* } \label{eq:covhatminusstar}\\
        &\quad\quad\quad\quad\quad\quad + \paren{{\tphistar{t} - \psi^* }}\paren{{\tphistar{t-i} - \psi^* }} - \ex{(\phistar - \psi^*)(\tphistar{t-i} - \psi^*)} \label{eq:covstarminusex}
    \end{align}
    To bound \cref{eq:covhatminusstar},
    \begin{align*}
        &T^{-1} \sum_{t=1}^T \sum_{i=1}^m \paren{\phihat - \psi(W_{1:T}; \hat \eta) } \paren{\tphihat{t-i} - \psi(W_{1:T}; \hat \eta) } - \paren{\phistar - \psi^* } \paren{\tphistar{t-i} - \psi^* } \\
        &= T^{-1} \sum_{t=1}^T\sum_{i=1}^m  \paren{\phihat - \psi(W_{1:T}; \hat \eta) - \phistar + \psi^* }\paren{\tphihat{t-i} - \psi(W_{1:T}; \hat \eta) } \\
        &\quad\quad + \paren{\phistar - \psi^* }\paren{\tphihat{t-i} - \psi(W_{1:T}; \hat \eta) - \tphistar{t-i} + \psi^* } \\
        &= T^{-1} \sum_{t=1}^T \paren{\phihat - \psi(W_{1:T}; \hat \eta) - \phistar + \psi^* }\sum_{i=1}^m \paren{\tphihat{t-i} - \psi(W_{1:T}; \hat \eta) } \\
        &\quad + T^{-1} \sum_{t=1}^T \paren{\phistar - \psi^* }\sum_{i=1}^m\paren{\tphihat{t-i} - \psi(W_{1:T}; \hat \eta) - \tphistar{t-i} + \psi^* } \\
        &= T^{-1} \sum_{t=1}^T o_{P}(1) \cdot \sum_{i=1}^m \paren{\tphihat{t-i} - \psi(W_{1:T}; \hat \eta) } \\
        &\quad + T^{-1} \sum_{t=1}^T \paren{\phistar - \psi^* }\sum_{i=1}^m o_{P}(1) 
    \end{align*}
    The first equality comes from the fact that for all $a,b,a',b'$, it holds $aa' - bb' = (a-b)a' + (a'-b')b$; the second comes from rearranging; the third comes from \cref{assm:2.2}. Now, we just need to argue that $\phistar - \psi^* = O_{P}(1)$ and $\tphihat{t} - \psi(W_{1:T}; \hat \eta) = O_{P}(1)$ for all $t$ to show that that \cref{eq:covhatminusstar} is $o_{P}(1)$.
    \begin{align*}
        \ex{(\phistar - \psi^*)^2} &\leq 2 \norm{\phistar}_{L^2(P)}^2 + 2 {\psi^*}^2 \leq C, \\
        \ex{(\tphihat{t} - \psi(W_{1:T}; \hat \eta))^2} &\leq 2\norm{\phihat}_{L^2(P)}^2 + 2 \norm{\psi(W_{1:T}; \hat \eta))}_{L^2(P)}^2 \leq C
    \end{align*}
    by \cref{assm:2.1} so Chebyshev's inequality completes the argument.
    \newcommand{\diffterm}[2]{\paren{\paren{{\tphistar{#1} - \psi^* }}\paren{{\tphistar{#2} - \psi^* }} - \ex{(\tphistar{#1} - \psi^*)(\tphistar{#2} - \psi^*)}}}
    \newcommand{\difftermnext}[2]{\paren{\paren{{\tphistar{#1} - \psi^* }}\paren{{\tphistar{#2} - \psi^* }} }}
    \newcommand{\difftermlast}[2]{\paren{\paren{{\tphistar{#1} - \psi^* }}\paren{{\tphistar{#2} - \psi^* }} }}
    To bound \cref{eq:covstarminusex},
    \begin{align*}
        &\ex{\paren{T^{-1} \sum_{t=1}^T\sum_{i=1}^m\paren{{\tphistar{t} - \psi^* }}\paren{{\tphistar{t-i} - \psi^* }} - \ex{(\phistar - \psi^*)(\tphistar{t-i} - \psi^*)}}^2} \\
        &= T^{-2}\sum_{t=1}^T\sum_{i=1}^m \expt{{\diffterm{t}{t-i}}^2} \\
        &\quad\quad\quad\quad+ 2  \sum_{j=1}^{m+i}\sum_{\ell=1}^{m+i}\mathbb{E}_{P}\bigg[\diffterm{t}{t-i} \\
        &\quad\quad\quad\quad\quad\quad\quad\quad\quad \cdot \diffterm{t-j}{t-\ell} \bigg] \tag{Rearranging; $m$-dependence.} \\
        &\leq T^{-2}\sum_{t=1}^T\sum_{i=1}^m \expt{{\difftermnext{t}{t-i}}^2} \\
        &\quad\quad\quad\quad+ 2  \sum_{j=1}^{m+i}\sum_{\ell=1}^{m+i}\bigg({\expt{\diffterm{t}{t-i}^2]}} \\
        &\quad\quad\quad\quad\quad\quad\quad\quad\quad \cdot {\expt{\diffterm{t-j}{t-\ell}^2}}\bigg)^{1/2} \tag{For r.v. $X$, $\E(X - \E(X))^2 \leq \E(X^2)$; Cauchy-Schwarz inequality.} \\
        &\leq T^{-2}\sum_{t=1}^T\sum_{i=1}^m \paren{\expt{\paren{{\tphistar{t} - \psi^* }}^4}\expt{\paren{{\tphistar{t-i} - \psi^* }}^4}}^{1/2} \\
        &\quad\quad\quad\quad+ 2  \sum_{j=1}^{m+i}\sum_{\ell=1}^{m+i}\bigg({\expt{\difftermnext{t}{t-i}^2]}}  \\
        &\quad\quad\quad\quad\quad\quad\quad\quad\quad \cdot {\expt{\difftermnext{t-j}{t-\ell}^2}}\bigg)^{1/2}\tag{\ditto}\\
        &\leq T^{-2}\sum_{t=1}^T\sum_{i=1}^m \paren{\expt{\paren{{\tphistar{t} - \psi^* }}^4}\expt{\paren{{\tphistar{t-i} - \psi^* }}^4}}^{1/2} \\
        &\quad\quad\quad\quad+ 2  \sum_{j=1}^{m+i}\sum_{\ell=1}^{m+i}\bigg({\expt{\paren{{\tphistar{t} - \psi^* }}^4}\expt{\paren{{\tphistar{t-i} - \psi^* }}^4}} \\
        &\quad\quad\quad\quad\quad\quad\quad\quad\quad \cdot {\expt{\paren{{\tphistar{t-j} - \psi^* }}^4}\expt{\paren{{\tphistar{t-\ell} - \psi^* }}^4}}\bigg)^{1/2} \tag{\ditto} \\
        &\leq T^{-2}\sum_{t=1}^T\sum_{i=1}^m \paren{\paren{8\norm{{\tphistar{t}}}_{L^4(P)}^4 + 8 {\psi^*}^4}{\paren{{\norm{8\tphistar{t-i}}_{L^4(P)}^4 + 8{\psi^*}^4 }}}}^{1/2} \\
        &\quad\quad\quad\quad+ 2  \sum_{j=1}^{m+i}\sum_{\ell=1}^{m+i}\paren{{\paren{{8\norm{\tphistar{t}}_{L^4(P)}^4 + 8{\psi^* }^4}}}{\bigg({{8\norm{\tphistar{t-i}}_{L^4(P)}^4 + 8{\psi^*}^4 }}}} \\
        &\quad\quad\quad\quad\quad\quad\quad\quad\quad \cdot {{\paren{{8\norm{\tphistar{t-j}}_{L^4(P)} + 8{\psi^*}^4 }}}\paren{{{8\norm{\tphistar{t-\ell}}_{L^4(P)}^4 + 8{\psi^*}^4 }}}}\bigg)^{1/2} \tag{for constants $a,b$, it holds $(a+b)^4 \leq 8a^4 + 8b^4$}\\
        &= o(1). \tag{\cref{assm:2.1}; there are $O(T)$ terms in the sum}
    \end{align*}
    Thus, with Chebyshev's inequality, \cref{eq:covstarminusex} is $o_{P}(1)$.
    Then, the fact that \cref{eq:covhatminusstar} is $o_{P}(1)$ and \cref{eq:covstarminusex} is $o_{P}(1)$ imply that \cref{eq:covterm2} is $o_{P}(1)$.
    Finally, the fact that \cref{eq:varterm} and \cref{eq:covterm2} are each $o_{P}(1)$ completes the proof of the lemma.
\qed
\fi

\section{Inference on Average Direct Effects} \label{sec:ade}

    In this section, we instantiate \cref{thm:main} to estimate average direct effects in observational settings with shared-state interference.
    The average direct effect (ADE), informally, is the mean difference between treatment and control outcomes for each individual, keeping all other individuals' treatment assignments the same.
    It is also sometimes called the expected average treatment effect \citep{savje_average_2019}.
    We first define a structural model of outcomes under interference in \cref{sub:ade}, apply our DML theorem to this setting in \cref{sub:ade_inf}, and then validate our method for finite samples via simulations in \cref{sub:ade_ni}.

    \subsection{Model, estimand and estimator} \label{sub:ade}

        Informally, our structural model will consist of data generated according to a standard observational setup for units indexed $t=1,\dots,T$ --- with an outcome $Y_t$, covariates $X_t$ and treatment assignment $D_t$ --- except that we will allow for each outcome to depend on a shared state $H_t$, which is dependent on data up to time $t-1$.
        The shared state, which may be vector-valued, will be the channel through which spillovers may occur in the data: outcomes are dependent on shared states and shared states may be dependent on previous outcomes, covariates, treatments and shared states.
        In particular cases, the shared state might represent available inventory in a market, like the number of drivers available on a ride-sharing platform, or public information, such as the popularity of songs on a music streaming platform.

        We will have, for each individual $t$, potential outcomes, shared-state and treatment given according to
        \begin{align}
            &Y_t(D_t, H_t) = f^*(D_t, X_t, H_t)  + \tilde Y_t, \label{eq:adey} \\
            &H_t(W_{t-1}) = h^*(W_{t-1}) + \tilde H_t , \label{eq:adeh} \\
            &D_t = m^*(X_t) + \tilde D_t, \label{eq:aded}
        \end{align}
        where $Y_t(\cdot, \cdot)$ is a potential outcome depending on unit $t$'s outcome and the shared state,
        $H_t(\cdot)$ is the potential shared state depending on the data from time $t-1$, $D_t$ is the treatment assignment and $X_t$ are covariates.
        We use $Y_t$ and $H_t$ to denote the random variables realized from the data generating process (as opposed to $Y_t(\cdot, \cdot)$ and $H_t(\cdot)$ which describe the response surface of \textit{potential} random variables under different treatment assignments and shared state or prior data).
        The structural model posits that $Y_t$ is determined by a function $f^*$ plus a stochastic residual term $\tilde Y_t$.
        Similarly, $H_t$ is a function of $h^*$ and residual $\tilde H_t$ and $D_t$ is a function of $m^*$ and residual $\tilde D_t$. 
        We will assume the residual terms $\tilde Y_t, \tilde H_t, \tilde D_t$ obey
        \begin{align}
            & \expt{\tilde Y_t \; | \; D_t, X_{t}, H_t} = 0, \nonumber \\ 
            & \expt{\tilde H_t \; | \; W_{t-1}} = 0, \nonumber \\
            &  \expt{\tilde D_t \; | \; X_{t}} = 0. 
        \end{align}
        Thus, $m^*$ is a propensity score function since $P(D_t = 1 \; | \; X_t) = \E_P(D_t \; | \; X_t) = m^*(X_t)$.

        By assumption, $H_t$ depends on data from prior time steps only through the observations at time $t-1$.
        Thus, $H_t$ satisfies the Markov property.
        We also note that this model allows for $H_t$ to be stochastic, even conditional on $W_{t-1}$, through the residual term $\tilde H_t$.
        Thus, the shared state may fluctuate due to factors exogenous to the previously arriving units' behavior.
        We also assume that $H_t$  satisfies either \cref{assm:markovchain}\cref{item:geoerg} or \cref{assm:markovchain}\cref{item:mdep}.
        This implies that the whole chain, consisting of observations $\{ W_t \}_{t=1}^T$ satisfies \cref{assm:markovchain}.

        Our estimand in this section will be the average direct effect (ADE).
        It is defined:
        \begin{align}
            \psi^* &\defeq \frac{1}{T} \sum_{t=1}^T \expt{Y_t(1, H_t) - Y_t(0, H_t)} \label{def:ade_psistar}
        \end{align}
        where the expectation is taken over $W_{1:T}$.
        In other words, this is the average effect on each individual of changing their treatment from control to treatment, without altering the distribution of others' treatments. (Note that the expectation $\E_P$ marginalizes over values of $H_t$, so the estimand is not conditional on realizations of $H_t$.)
        
        Recall from \cref{sec:dml4ssi} that, in our setting the estimator $\psi(W_{1:T};\eta)$ is the sample average of functions $\varphi(W_t; \eta)$. 
        In this section (and the next), the functional form of the time-$t$ estimator $\varphi$ can be written as the sum of two components: a plug-in term and a debiasing term.
        For convenience, we will denote these $\plugin$ and $\debiasing$; i.e.,
        $\varphi(W_t; \eta) = \plugin(W_t; \eta) + \debiasing(W_t; \eta)$.
        These will be defined in this section as
        \begin{align}
            \begin{aligned}
            &\plugin(W_t, \eta) \defeq f(1, X_t, H_t) - f(0, X_t, H_t), \\
            &
            \ificml
\else
                \debiasing(W_t, \eta) \defeq \paren{\frac{D_t}{m(X_t)} - \frac{1 - D_t}{1 - m(X_t)}} 
                 \cdot (Y_t - f(D_t, X_t, H_t)),
            \fi
            \end{aligned} \label{def:ade_psi}
        \end{align}
        where the nuisance estimators are $\eta = (f, m)$.
        The function $\varphi$ mirrors the augmented inverse probability weighting (AIPW) estimator \citep{robins_estimation_1994} that is standard in the semiparametric inference literature (see, e.g., \citet{wager_causal_2024} for background and history), with the exception that our outcome nuisance estimator $f$ takes the shared state $H_t$ as an argument. 
        One way to interpret our estimator is as a covariate adjustment for the shared state: omitting the shared-state at a given time may result in a form of confounding between units' outcomes.
        However, it is not sufficient to treat $H_t$ as a covariate and apply DML methods as if the data were iid, since there is dependence between units which needs to be accounted for.
        We will see the contrast between our method and an approach that treats the shared state as if they were iid covariates in our simulations, and we note that, in \cref{sec:sb}, treating the shared state as if they are covariates will not in general yield a consistent treatment effect estimator.

    \subsection{Estimation and inference} \label{sub:ade_inf}

        Before we state our main result in this section, we formalize assumptions sufficient for the result to hold.

        %
        In \cref{assm:ade_regularity}, we require that each individual has true and predicted probability of assignment to treatment and to control bounded away from zero, almost surely. 
        This is standard in settings that depend on estimation of the propensity score, since otherwise treatment assignments that occur with small probability may have a large influence on the estimator, making it unstable even as the sample size grows.
        We also require that the outcome function be bounded in $L^q$ norm, for $q > 4$ and that the error terms $\tilde Y_t$ are bounded in $L^2$ norm. 
        These ensure that the concentration properties of the data $W_{1:T}$ propagate to the estimator. We state it formally next.
        \begin{assumption}[Regularity conditions for ADE estimation] \label{assm:ade_regularity}
            There exist constants $C, \zeta  > 0$ such that the following regularity conditions are met for all $(f, m) \in S_T$ and all $t \in [T]$
            \begin{align}
                &\zeta < m(X_t) < 1 - \zeta, \text{ a.s.}  \label{assm:boundedmhat}\\
                &\norm{f(D_t,X_t,H_t)}_{L^{4 + \delta}(P)} < C,  \label{assm:f_reg}\\ 
                &\| {\tilde Y_t} \|_{L^2(P)} < C  \label{assm:regularerror}
            \end{align}
        \end{assumption}
        We note that since $m^*, f^* \in S_T$ by assumption, the regularity conditions \cref{assm:boundedmhat,assm:f_reg} also impose restrictions on the true nuisances.

        In \cref{assm:ade_rates}, we require that the nuisance estimators $\hat m$ and $\hat f$ converge in $L^2$ to their true values $m^*$ and $f^*$ and that the product of their convergence rates vanishes at $o_P({T}^{-1/2})$ rates.
        Recall that for $q \geq 1$ the $L^q$ norm of a random variable $Z$ is $\E[|Z|^q]^{1/q}$.
        \begin{assumption}[Rate conditions for ADE estimation]\label{assm:ade_rates}
            The following rate conditions are met for all $(f, m) \in S_T$
            \begin{align}
                &\frac{1}{T} \sum_{t=1}^T\norm{(m^* - m)(X_t)}_{L^2(P)} = o(1), \label{assm:l2mminusmstar}\\
                &\frac{1}{T} \sum_{t=1}^T\norm{(f^* - f)(D_t, X_t, H_t)}_{L^2(P)} = o(1), \label{assm:l2fminusfstar}\\ 
                &\bigg| { T^{-1} \sum_{t=1}^T \expt{(m-m^*)(X_t) \cdot (f-f^*)(D_t, X_t, H_t)}} \bigg| = o_{P_T}(T^{-1/2}),\label{assm:productrate}
            \end{align}
        \end{assumption}
        The assumption is quantified over all $(f,m)$ in the nuissance realization set $S_T$.
        \Cref{assm:l2mminusmstar} says that, if on average over units $t=1,\dots,T$, the expected squared difference between $m$ and $m^*$ must go to zero.
        \Cref{assm:l2fminusfstar} says the same for $f$ and $f^*$.
        \Cref{assm:productrate} says that the average of the product in $m-m^*$ and $f-f^*$ over the data must go to zero at faster-than-$\sqrt{T}$ rates.
        
        The required rates for \Cref{assm:ade_rates} can be achieved if, for example, when averaging over time steps, $\norm{(m - m^*)(X_t)}_{L^2(P)} = o(T^{-1/4})$ and $\norm{(f - f^*)(D_t, X_t, H_t)}_{L^2(P)} = o(T^{-1/4})$.
        On the other hand, if the propensity score function $m^*$ is known (if, e.g., it is determined via a randomized experiment controlled by the researcher), then $\norm{(f - f^*)(D_t, X_t, H_t)}_{L^2(P)}$ can converge at an arbitrarily slow rate and still yield an efficient estimator.

        There is a rich literature giving learning rates for machine learning estimators trained on dependent data.
        A key implication of our Markov chain assumptions is that the data are $\rho$-mixing, which intuitively means that observations that are far apart have diminishing correlations (see, e.g., \cite{bradley_chapter_2007} for formal definitions of mixing conditions).
        Learning rates for mixing sequences have been proved for neural networks \citep{ma_theoretical_2022}, random forests \citep{goehry_random_2020} and several other machine learning methods \citep{irle_consistency_1997,steinwart_learning_2009,lozano_convergence_2014,wong_lasso_2020}.
        See \cite{ballinari_semiparametric_2024} for further discussion. 

        With these two assumption in hand, we are ready to state our efficient inference result for this section.

        \begin{theorem} \label{thm:ade}
            For the model defined in \cref{sub:ade} and the estimand $\psi$ and estimator $\psi^*$ defined in \cref{def:ade_psi,def:ade_psistar} respectively, under \cref{assm:ade_regularity,assm:ade_rates},  
            then, with probability no less than $1 - \gamma$, it holds $$\sqrt{T}\hat\sigma^{-1}( \psi(W_{1:T}; \hat\eta) - \psi^*) \convindist N(0, 1)$$ where $\hat\sigma$ is defined in \cref{eq:sigmasqhatdef}.
        \end{theorem}

        \Cref{thm:ade} instantiates \Cref{thm:main} for the model defined in \Cref{sub:ade}.
        It replaces the general assumptions required for \Cref{thm:main} with specific regularity and rates assumptions in \Cref{assm:ade_regularity,assm:ade_rates}.

        \ifproofsinbody
            \noindent \textit{Proof of \cref{thm:ade}.}
If we can verify that \cref{assm:markovchain,assm:regularity,assm:rates} hold, then we can apply \cref{thm:main}.
Recall that \cref{assm:markovchain} is verified by assumption in the definition of the model in \cref{sub:ade}. 
\Cref{assm:regularity}\cref{item:twicegd} is trivially verified from the definition of $\psi$.
\Cref{assm:regularity}\cref{item:neymanorth} is given by \cref{lem:ade_neymanorth}.
\Cref{assm:rates}, \cref{assm:2.1} is trivially verified by the definition of $\psi$ and \Cref{assm:boundedmhat,assm:f_reg}.
\Cref{assm:rates}, \cref{assm:2.2} is given by \Cref{lem:ade_consistency}.
\Cref{assm:rates}, \cref{assm:2.3} is given by \Cref{lem:ade_2ndorder}.
\qed

\vspace{1em}

\Cref{lem:ade_neymanorth} states that the estimator $\psi$ must be Neyman orthogonal with respect to the nuisance realization set $S_T$.
The argument is standard to any proof of Neyman orthogonality of an augmented inverse probability weighted (AIPW) estimator.

\begin{lemma}[Neyman orthogonality] \label{lem:ade_neymanorth}
For the model defined in \cref{sub:ade} and the estimand $\psi$ and estimator $\psi^*$ defined in \cref{def:ade_psi,def:ade_psistar} respectively, and under \cref{assm:ade_regularity,assm:ade_rates}, $\psi(W_{1:T}; \eta)$ is Neyman orthogonal with respect to $S_T$.
\end{lemma}

\noindent \textit{Proof.}
    Notice, for all $t$
    \begin{align*}
        &\frac{\partial}{\partial r} \expt{\varphi(w_{t}; \eta^* + r(\eta - \eta^*)} \bigg|_{r=0} \\
        &= \E \bigg[(f-f^*)(1, X_t, H_t) - (f-f^*)(0, X_t, H_t) \\
        &\quad + \paren{\frac{D_t}{m^*(X_t)} - \frac{1 - D_t}{1 - m^*(X_t)}}(f-f^*)(D_t, X_t, H_t) \bigg] \tag{\cref{assm:f_reg} and the dominated convergence theorem}\\
        &= \expt{(f-f^*)(1, X_t, H_t) \paren{1 - \frac{D_t}{m^*(X_t)}} - (f-f^*)(0, X_t, H_t) \paren{1 - \frac{1 - D_t}{1 - m^*(X_t)}}} \tag{Rearranging}\\
        &= 0. \tag{$\expt{ D_t / m^*(X_t) \; | \; X_t} = \expt{(1-D_t) / (1 - m^*(X_t))\; | \; X_t} = 1$}
    \end{align*}
\qed

\vspace{1em}

\Cref{lem:ade_consistency} states that, on average over time, $\varphi(W_t; \eta)$ must converge to $\varphi(W_t; \eta^*)$ in $L^2$, uniformly over the nuisance realization set $S_T$.
The proof consists of showing that convergence of $\varphi$ reduces to conditions on convergence of $f$ to $f^*$ and $m$ to $m^*$, which we assumed in \cref{assm:ade_rates}.

\begin{lemma}[Consistency] \label{lem:ade_consistency} 
For the model defined in \cref{sub:ade} and the estimand $\psi$ and estimator $\psi^*$ defined in \cref{def:ade_psi,def:ade_psistar} respectively, under \cref{assm:ade_regularity,assm:ade_rates}, 
\begin{align*}
    \sup_{\eta \in S_T} T^{-1} \sum_{t=1}^T \norm{ \varphi(W_{t}; \eta) - \varphi(W_{t}; \eta^*)}_{L^2(P)} = o(1).
\end{align*}
\end{lemma}

\noindent \textit{Proof.} 
Notice, for $\eta \in S_T$,
\begin{align*}
    &\varphi(W_{t}; \eta) - \varphi(W_{t}; \eta^*) \\ &= (f-f^*)(1, X_t, H_t) - (f - f^*)(0, X_t, H_t) \\
    &\quad + D_t\paren{\frac{Y_t - f(D_t, X_t, H_t)}{m(X_t)} - \frac{Y_t - f^*(D_t, X_t, H_t)}{m^*(X_t)} } \\
    &\quad - (1 - D_t)\paren{\frac{Y_t - f(D_t, X_t, H_t)}{1 - m(X_t)} - \frac{Y_t - f^*(D_t, X_t, H_t)}{1 - m^*(X_t)}}  \\
    &= (f-f^*)(1, X_t, H_t) - (f - f^*)(0, X_t, H_t) \\
    &\quad + D_t\paren{\frac{(f^* - f)(D_t, X_t, H_t) + \tilde Y_t}{m(X_t)} - \frac{\tilde Y_t}{m^*(X_t)} } \\
    &\quad - (1 - D_t)\paren{\frac{(f^* - f)(D_t, X_t, H_t) + \tilde Y_t}{1 - m(X_t)} - \frac{\tilde Y_t}{1 - m^*(X_t)}} \tag{\cref{eq:adey}}\\
    &\leq (f-f^*)(1, X_t, H_t) - (f - f^*)(0, X_t, H_t) \\
    &\quad\quad + \zeta^{-2} D_t \paren{\paren{(f-f^*)(D_t, X_t, H_t) }m^*(X_t) - \tilde Y_t ((m^* - m)(X_t))}  \\
    &\quad\quad + \zeta^{-2} (1 - D_t) \paren{\paren{(f-f^*)(D_t, X_t, H_t) }(1-m^*(X_t)) - \tilde Y_t ((m - m^*)(X_t))}  \tag{\cref{assm:boundedmhat}}\\
    &\leq (f-f^*)(1, X_t, H_t) - (f - f^*)(0, X_t, H_t) \\
    &\quad\quad + 2\zeta^{-2} \paren{\paren{(f-f^*)(D_t, X_t, H_t) } - \tilde Y_t ((m - m^*)(X_t))}  \tag{$D_t, 1-D_t, m^*(X_t), 1 - m^*(X_t)$ are all in $[0,1]$ } 
\end{align*}
Thus, by the triangle inequality:
\begin{align*}
    &\sum_{t=1}^T \norm{ \varphi(W_{t}; \eta) - \varphi(W_{t}; \eta^*)}_{L^2(P)}\\
    &= \frac{1}{T} \sum_{t=1}^T \norm{(f-f^*)(1, X_t, H_t)}_{L^2(P)} \\
    &\quad\quad\quad+ \norm{(f-f^*)(1, X_t, H_t)}_{L^2(P)} \\
    &\quad\quad\quad+ 2\zeta^{-2} \paren{\norm{(f-f^*)(D_t, X_t, H_t) }_{L^2(P)} + O_P(1) \norm{m(X_t) - m^*(X_t)}_{L^2(P)}} \tag{\cref{assm:regularerror}}
\end{align*}
Next, the assumptions \cref{assm:boundedmhat} and \cref{assm:l2fminusfstar} combined imply
\begin{align*}
    \frac{1}{T} \sum_{t=1}^T\norm{(f-f^*)(1, X_t, H_t)}_{L^2(P)} = o(1) \\
    \frac{1}{T} \sum_{t=1}^T\norm{(f-f^*)(0, X_t, H_t)}_{L^2(P)} = o(1) .
\end{align*}
And finally
\begin{align*}
    2 \zeta^{-2} \frac{1}{T} \sum_{t=1}^T \norm{(f-f^*)(D_t, X_t, H_t)}_{L^2(P)} = o(1) \\
    2 \zeta^{-2} \frac{1}{T} \sum_{t=1}^T O_P(1) \norm{(m-m^*)(X_t)}_{L^2(P)} = o(1) 
\end{align*}
by \cref{assm:l2fminusfstar,assm:l2mminusmstar}. \qed

\vspace{1em}

\Cref{lem:ade_2ndorder} states that the second-order term in Taylor's theorem converges to zero at faster-than-$\sqrt{T}$ rates.
This ensures we can apply Taylor's theorem and establishes that slow rates of convergence of $\hat\eta$ to $\eta^*$ do not lead to sub-optimal rates on convergence of $\psi(W_{1:T}, \hat\eta)$ to $\psi^*$.
The proof relies on showing that the second Gateaux derivative of $\psi$ on the path from any $\eta$ in the nuisance realization set $S_T$ to $\eta^*$ reduces to conditions on the \textit{products} of convergence rates of $m$ to $m^*$ and $f$ to $f^*$.
Thus, it is sufficient for each of $m$ and $f$ to converge to their true values at faster-than-$T^{1/4}$ rates, or for $m$ to be known \textit{a priori} while $f$ converges at arbitrarily slow rates.

\begin{lemma}[Second-order condition] \label{lem:ade_2ndorder}
For the model defined in \cref{sub:ade} and the estimand $\psi$ and estimator $\psi^*$ defined in \cref{def:ade_psi,def:ade_psistar} respectively, under \cref{assm:ade_regularity,assm:ade_rates},
\begin{align*}
    \sup_{r \in(0,1), \eta \in S_T}\abs{ \frac{\partial^{2}}{\partial r^{2} } \expt{ \psi(W_{1:T}; \eta^* + r(\eta - \eta^*))}} = o({T^{-1/2}}).
\end{align*}
    
\end{lemma}
\noindent \textit{Proof.} Observe, for $\eta \in S_T$ and $r \in (0, 1)$,
    \begin{align*}
        &\frac{\partial^2}{\partial r^2} \expt{\varphi(w_{t}; \eta^* + r(\eta - \eta^*)} \\
        &= \E_P \bigg[ {- \paren{\frac{D_t}{(m^*(X_t) + r((m-m^*)(X_t)))^2} - \frac{1 - D_t}{(1 - m^*(X_t)- r((m-m^*)(X_t)))^2}}}  \\
        &\quad\quad\quad \cdot (m-m^*) (X_t)\cdot  (f-f^*)(D_t, X_t, H_t)\bigg]
    \end{align*}
    where we first switch the derivative with the expectation using \cref{assm:f_reg} and the dominated convergence theorem, and then we evaluate the derivatives.
    This implies
    \begin{align*}
         &\abs{\frac{\partial^{2}}{\partial r^{2} } \expt{ \psi(W_{1:T}; \eta^* + r(\eta - \eta^*))}} \\
         &= \bigg| T^{-1} \sum_{t=1}^T \E_P \bigg[ {{\frac{D_t}{(m^*(X_t) + r((m-m^*)(X_t)))^2} - \frac{1 - D_t}{(1 - m^*(X_t)- r((m-m^*)(X_t)))^2}}} \\
         &\quad\quad\quad\quad\quad\quad\quad\quad \cdot {(m-m^*) (X_t)\cdot  (f-f^*)(D_t, X_t, H_t)} \bigg] \bigg| \\
         &\leq \abs{ T^{-1} \sum_{t=1}^T \E [\zeta^{-2}  \cdot {(m-m^*) (X_t)\cdot  (f-f^*)(D_t, X_t, H_t)}] } \tag{\cref{assm:boundedmhat}} \\
         &\leq o_P(T^{-1/2}). \tag{\cref{assm:productrate}}
    \end{align*}
\qed

        \fi
        
    \subsection{Simulations and estimator comparison} \label{sub:ade_ni}

    Next, we validate our results through simulations demonstrating the performance of the double machine learning estimator versus naive alternate estimators.

    \paragraph{Setup for treatment effect estimator comparisons.} We compare our estimator, which we will sometimes call the DML4SSI estimator, to a naive plug-in estimator $\psi^{\mathrm{pi}}$, defined as \begin{align}
         \psi^{\mathrm{pi}}(W_{1:T}; \eta) \defeq T^{-1} \sum_{t=1}\varphi^{\mathrm{pi}}(W_{t}; \eta) \label{eq:ade_plugin_psi}
     \end{align} where $\varphi^{\mathrm{pi}}$ is defined as in \cref{def:ade_psi}.
     Plug-in estimators constructed using machine learning models will converge at slower-than-$\sqrt{T}$ rates, and in finite samples may be very biased.
     We also compare our results to a naive difference-in-means Horvitz-Thompson (HT) estimator $\psi^\mathrm{HT}$, defined as 
     \begin{align}
         \psi^\mathrm{HT} (W_{1:T}; \eta) \defeq T^{-1} \sum_{t=1}^T \paren{\frac{D_t}{m(X_t)} - \frac{1-D_t}{1-m(X_t)}} Y_t. \label{eq:naive_ht}
     \end{align}
     The naive HT estimator does not account for the shared state and thus may be inconsistent and very biased in finite samples.
     Finally, we include a comparison with a naive DML estimator (DML-N) which treats the data as if it were iid and fails to account for the shared state.
     The naive DML estimator $\psi^{\mathrm{DML-N}}$ is defined as the canonical iid AIPW estimator, ignoring the shared state:
     \begin{align}
       \psi^{\mathrm{DML-N}}(W_{1:T}; \eta') \defeq T^{-1} \sum_{t=1}^T \varphi^{\mathrm{DML-N}}(W_{t}; \eta') \label{eq:dmlnaive}
     \end{align}
     where
     \begin{align*}
         \varphi^{\mathrm{DML-N}}(W_{t}; \eta') \defeq f'(1, X_t) - f'(0, X_t) + \paren{\frac{D_t}{\hat m(X_t)} - \frac{1 - D_t}{1 - \hat m(X_t)}} \cdot (Y_t - f'(D_t, X_t)).
     \end{align*}
     The nuisances are $\eta' = (f', \hat m)$, where we learn an outcome model $f'$ by learning a predictor of $Y_t$ given $X_t$ and $D_t$ but omitting the shared state, and learn $\hat m$ as before by predicting $D_t$ as a function of $X_t$.
     (We use the notation $f'$ to disambiguate from $\hat f$, which takes $H_t$ as an argument.)
     To learn $f'$, we use an auxiliary sample as in \Cref{alg:dmlssv}.
     Like the HT estimator, $\psi^{\mathrm{DML-N}}$ does not account for influence of the shared state and thus may in general be inconsistent and very biased in finite samples.

\begin{figure}[t]
         \centering
         \ificml
\else
        \includegraphics[width=0.7\columnwidth]{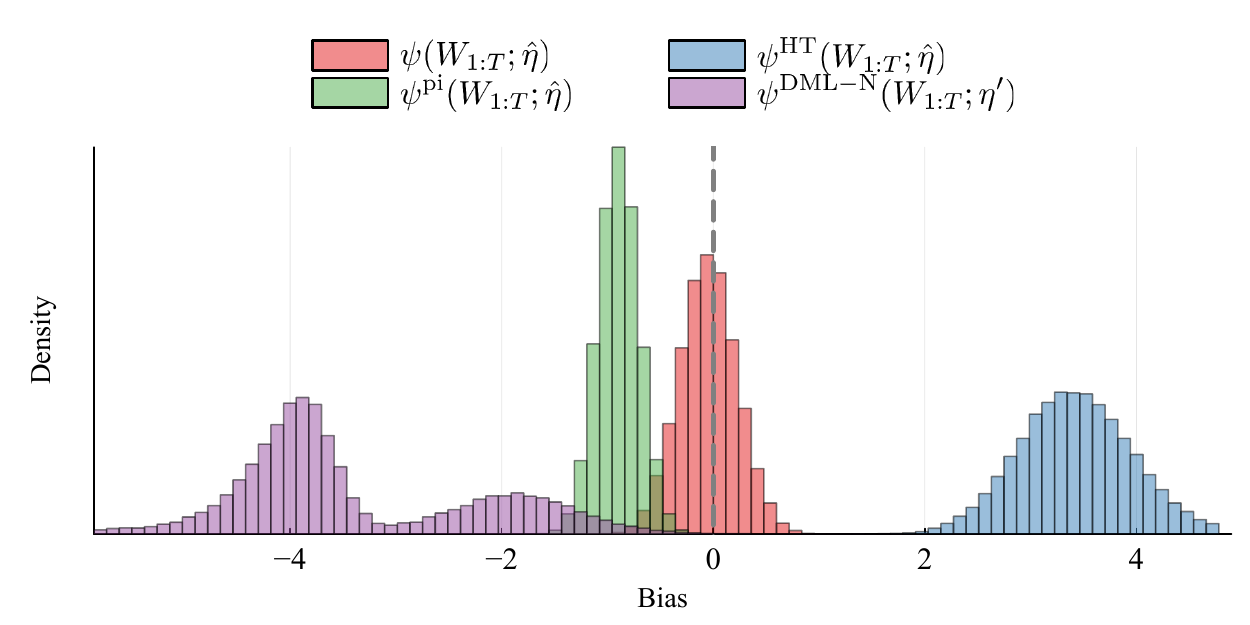}
        \fi
         \caption{Estimates of the average direct effect with our double machine learning estimator versus naive Horvitz-Thompson, naive plug-in and naive DML estimators.}
         \label{fig:ade}
     \end{figure}

    In our simulations, we generated data according to a smooth function of both $D_t, X_t$ and of $H_t$.
    %
    %
    %
    For all machine learning predictors $\hat f$, $\hat m$ and $f'$, we used random forests with default parameters trained on auxiliary data of size $T$ sampled independently of the data used for inference.
    Then, we computed each of the estimators $\psi(W_{1:T}; \hat \eta), \psi^{\mathrm{HT}}(W_{1:T}; \hat \eta)$, $\psi^{\mathrm{pi}}(W_{1:T}; \hat \eta)$ and $\psi^{\mathrm{DML-N}}(W_{1:T}; \eta')$ and created a density plot for the bias of the resulting estimates in \cref{fig:ade}.
    Full details of the simulation are available in \cref{sec:simulationdetails} and the code used to generate them is available at \href{https://github.com/johnchrishays/dml4ssi}{\texttt{https://github.com/johnchrishays/dml4ssi}}.

    \paragraph{Results for treatment effect estimator comparisons.} In \cref{fig:ade}, we observe that all the distributions of each of the naive estimators are substantially biased.
    By contrast, the DML estimator is centered around the true treatment effect.
    %
    The bias in the plug-in estimator occurs despite the smoothness of $f^*$ and $m^*$ and the fact that the data is low dimensional. The bias in the HT and DML-N estimators comes from the fact that they do not account for the effect of the shared state and thus will in general be inconsistent.
    The direction and magnitude of the bias for each naive estimator is idiosyncratic to our synthetic data generating process and machine learners; in other settings the sign of the biases may be reversed, and the relative performance of different naive estimators may change.

     \paragraph{Setup for confidence interval comparisons.} We next investigate the coverage of confidence intervals constructed from treatment effect and variance estimators.
     Confidence intervals for an arbitrary pair of treatment effect, variance estimators $(\hat \psi, \hat \sigma^2)$ are constructed as 
     \newcommand{\CI}{\mathrm{CI}}
     \begin{align*}
         \CI_\alpha(\hat \psi, \hat \sigma^2) = [\hat \psi - z_\alpha \hat \sigma / \sqrt{T}, \hat \psi + z_\alpha \hat \sigma / \sqrt{T}]
     \end{align*}
     where $z_\alpha$ is the $(1-\alpha/2)$-th quantile of a standard normal distribution.
     To compute coverage rates, we run many simulations, constructing a confidence interval for each one and calculating the proportion of confidence intervals that contain the estimand $\psi^*$.
     
     The DML4SSI confidence intervals are constructed using $\psi(W_{1:T}, \hat\eta)$ and the consistent variance estimator $\hat \sigma^2$ in \Cref{thm:main}.
     The Horvitz-Thompson (HT) confidence intervals use the treatment effect estimator $\psi^{\mathrm{HT}}(W_{1:T}; \hat \eta)$ as defined in \cref{eq:naive_ht} and a standard variance estimator $\hat\sigma^2_{\mathrm{HT}}$ defined as
     \begin{align*}
         \hat\sigma^2_{\mathrm{HT}} &= T^{-1} \sum_{t=1}^T \paren{\frac{D_t}{\hat m(X_t)} (Y_t - \overline{Y}^{(0)}) - \frac{1-D_t}{1-\hat m(X_t)} (Y_t - \overline{Y}^{(1)})}^2
     \end{align*}
     where $\overline{Y}^{(\ell)}$ for $\ell \in \{0, 1\}$ is the mean observed outcome conditional on $D_t = \ell$. 
     The plug-in confidence intervals use $\psi^{\mathrm{pi}}(W_{1:T}; \eta)$ as defined in \cref{eq:ade_plugin_psi} and the variance estimator defined in \cref{thm:main}, substituting $\varphi^{\mathrm{pi}}$ for $\varphi$ and $\psi^{\mathrm{pi}}$ for $\psi$.
     The naive DML (DML-N) estimator uses the treatment effect estimator $\psi^{\mathrm{DML-N}}$ defined as
     \begin{align}
         \hat\sigma^2_{\mathrm{DML-N}} = T^{-1} \sum_{t=1}^T \paren{\varphi^{\mathrm{DML-N}}(W_t; \hat \eta) - \psi^{\mathrm{DML-N}}(W_{1:T}; \hat \eta)}^2. \label{eq:ssacvar}
     \end{align}

     In addition to the HT, plug-in and DML-N estimators compared above, we also compare our method to one where we treat the shared-state as covariates and construct treatment effect and variance estimates accordingly.
     We call these the \textit{shared-state as covariates (SSAC)} estimators.
     In this section, the SSAC treatment effect estimator is mechanically the same as our treatment effect estimator, since both of them involve learning a conditional expectation function estimate $\hat f$.
     (In \cref{sec:sb}, this is not true, and our treatment effect estimator and the SSAC estimators will be different, leading to inconsistent SSAC treatment effect estimates.) 
     However, the SSAC variance estimator ignores covariances between terms and treats the data as if there were no covariances between terms. This is different from our variance estimator, which does account for covariances between terms.
     The SSAC variance estimator $\hat\sigma^2_{\mathrm{SSAC}}$ is defined as in \cref{eq:ssacvar}, substituting $\varphi$ for $\varphi^{\mathrm{DML-N}}$ and $\psi$ for $\psi^{\mathrm{DML-N}}$.
     These are the variance estimates that would have been constructed had the shared states been treated like iid covariates.

\begin{figure}[t]
         \centering
         \begin{subfigure}{0.49\textwidth}
             \centering
             \includegraphics[width=\linewidth]{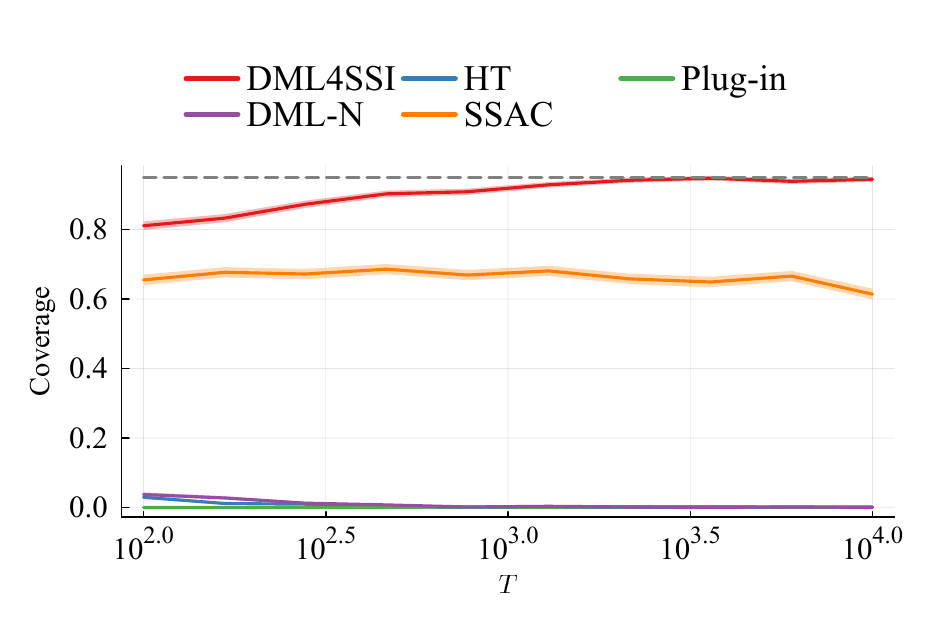}
         \end{subfigure}
         \begin{subfigure}{0.49\textwidth}
             \centering
             \includegraphics[width=\linewidth]{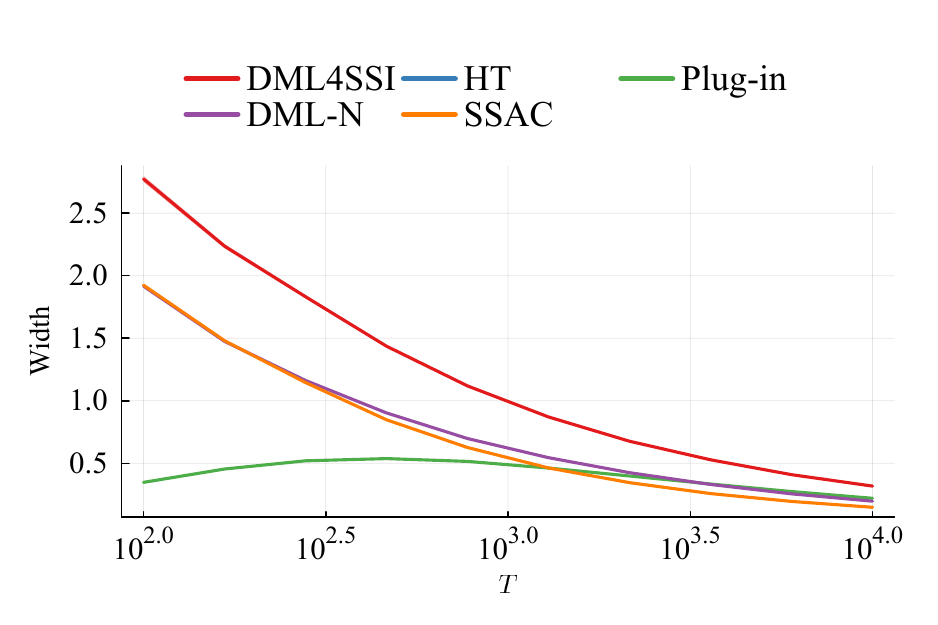}
         \end{subfigure}
         \caption{Coverage rates of 95\% confidence intervals for the ADE constructed using our estimators (DML4SSI), shared state as covariates (SSAC) estimators, naive Horvitz-Thompson estimators, plug-in estimators and naive DML estimators. The HT, plug-in and naive DML estimators coverage overlap around 0. We plot standard errors of the simulations in lighter-colored ribbons around the coverage point estimates.}
         \label{fig:coverage}
     \end{figure}

     \paragraph{Results for confidence interval comparisons.} In \cref{fig:coverage}, we run simulations for different values of $T$ and construct confidence intervals for each estimator.
     In the left plot, we show coverage rates of the confidence intervals, and in the right plot, we show the confidence interval widths.
     We observe that the DML4SSI confidence intervals approach the target 95\% coverage as $T$ increases, while each of the other estimators fall substantially below the target coverage.
     In particular, each of the HT, plug-in and DML-N confidence are substantially biased as we saw in \cref{fig:ade}, leading to confidence intervals with coverage close to zero.
     %
     The SSAC confidence intervals use the same (consistent) treatment effect estimator as the DML4SSI but have variance estimates that do not account for covariance between observations over time.
     This leads to underestimates of the variance, confidence intervals that are too narrow and lower-than-desired coverage.
     In this case, the SSAC 95\% confidence intervals result in coverage between 0.6 and 0.7, and do not seem to be converging to the target coverage as $T$ increases.

\section{Regression adjustments in switchback experiments} \label{sec:sb}

    In this section, we turn our attention to estimating the global average treatment effect (GATE) using switchback experiments.
    The GATE, as we will define formally shortly, is the difference in an the mean outcome of interest when all units are assigned to treatment versus when all units are assigned to control.
    Switchback experiments, which were formalized in \citet{bojinov_design_2022}, allow for valid inference in settings where interference is $m$-dependent.
    %
    %

    Switchback experiments are implemented by assigning treatments in sequential blocks so that $\ell > m$ units in a row receive the same (randomized) treatment.
    For example, for a block of $\ell$ sequential units, the experimenter may use a Bernoulli random draw to assign all units in the block to treatment or to control.
    Intuitively, switchback experiments account for interference by arranging for the observation of some units where all $m$ previous units received the same treatment assignment as the current unit.
    Our framework incorporates doubly robust regression adjustments that allow for lower variance estimates in settings where covariates and shared-state variable exert significant influence on outcomes (or, equivalently, in cases where treatment effect sizes are small).
    The structure of the remainder of this section is the same is in \cref{sec:ade}: we define our setting in \cref{sub:sb_model}, state an efficient inference result in \cref{sub:sb_inf}, and validate our results for finite samples via simulations in \cref{sub:sb_ni}.

    \subsection{Model, estimand and estimator} \label{sub:sb_model}

    Our model in this section is similar to that of \cref{sub:ade}: for $t=1,\dots,T$, we have an outcome $Y_t$ that depends on a shared state $H_t$, covariates $X_t$ and treatment assignment $D_t$.
    This section is different from the previous in that we require $H_t$ to consist of the last $m$ observations of treatment assignments and covariates.
    This allows for satisfying the $m$-dependence condition required for switchback experiments to yield unbiased estimates of treatment effects \citep{bojinov_design_2022}.
    Also, since we are considering switchback experiments, we assume $D_{1:T}$ to come from a known distribution determined by the experimenter.

    Formally, we will analyze the following potential outcomes model on a population indexed $t=1,2,\dots$. For a constant $m \in \N$ and for each individual $t$ there is some binary treatment of interest $D_t \in \{ 0, 1 \}$, iid covariates $X_t \in \R^{d}$, shared state $H_t \in \R^{d_H}$ and outcome $Y_t \in \R$ given by
    \begin{align}
        &Y_t(D_t, H_t) = f^*(D_t, X_t, H_t) + \tilde Y_t,   \label{eq:sby} \\
        &H_t(D_{(t-m):(t-1)}) 
        = \textvec(D_{(t-m):(t-1)}, X_{(t-m):(t-1)}) \\
        &D_{1:T} \sim \cD
    \end{align}
    where $\cD$ is a distribution determined by the experimenter. We will drop the arguments to $H_t$ when $D_{(t-m):(t-1)}$ are the observed treatment assignments. The residual terms $\tilde Y_t$ will be assumed to be mean zero and obey
    \begin{align}
        \expt{\tilde Y_t \; \big| \; D_t, X_t, H_t} = 0.
    \end{align}
    Notice that the above implies that shared states are $m$-dependent as defined in \cref{def:mdep}.
    For an index set $\mathbf{t} = (s:t)$, $s,t \in [T]$ and boolean vector $\mathbf{b} = \{ 0, 1 \}^{|\mathbf{t}|}$, define
    \begin{align}
        \pi^*(\mathbf{t}, \mathbf{b})&\defeq \prob{ D_{\mathbf{t}} = \mathbf{b}}   \label{eq:sbd}
    \end{align}
         \label{sub:switchbackestimator}
    to be the probability that the subvector $D_\mathbf{t}$ is equal to $\mathbf{b}$.
     The estimand will be the GATE, defined as
     \begin{align}
         \psi^* = \frac{1}{T} \sum_{t=1}^T \expt{Y_t(1, H_t(\1)) - Y_t(0, H_t(\0))}. \label{def:sb_psistar}
     \end{align}
     As in the previous section, our estimator $\psi$ will be defined by the sample average of the sum of a plug-in term $\plugin(W_t, \eta)$ and a debiasing term $\debiasing(W_t, \eta)$. We define these terms for this section as follows:
     \begin{align}
     \begin{aligned}
            &\plugin(W_t, \eta) \defeq f(1, X_t, H_t(\1)) - f(0, X_t, H_t(\0)),  \\
            \ificml
\else
            &\debiasing(W_t, \eta) \defeq \paren{\frac{\indic{\dvec = \1}}{\pi^*({(t-m):t}, \1)} - \frac{\indic{\dvec = \0}}{\pi^*({(t-m):t}, \0)}} 
             \paren{Y_t - f(D_t, X_t, H_t)}.
            \fi
     \end{aligned}
        \label{def:sb_psi} 
     \end{align}

     Note that the estimator has a form similar to the AIPW estimator of the previous section with a few differences. 
     First, in the plug-in component of the estimator, we plug in the all treatment or all control vector for treatment assignments.
     Second, in the debiasing component of the estimator, we substitute in indicator functions for the $m$ past treatment assignments being equal to the current units' treatment assignment.
     These ensure that the estimator identifies the effect of all treatment versus all control.

    {We will assume $\pi^*({(t-m):t}, \1), \pi^*({(t-m):t}, \0) > \zeta$ for all $t$. This is not always the way that switchback experiments are defined: in some contexts, the switching points may be deterministic, which would mean that any $t$ less than $m$ time steps after a (deterministic) switch point would have $\pi^*({(t-m):t}, \1)= \pi^*({(t-m):t}, \0) = 0$.
    One way to achieve the constraint is to pick a switching length $\ell > m$ and then pick a first switch uniformly at random from the first $\ell$ time steps. Then for each block of constant treatment assignments, Bernoulli randomize. This will give $P(D_{(t-m):t} = \mathbf{1}) = P(D_{(t-m):t} = \mathbf{0}) \geq (\ell-m)/2\ell$.}

    \subsection{Estimation and inference} \label{sub:sb_inf}

    The assumptions necessary for efficient inference in switchback experiments under $m$-depen\-dence are analogous to those in \cref{sec:ade} and are deferred to the appendix. In \cref{assm:sb_regularity}, we require that the probability that any $m+1$ sequential treatment assignments are the same (either all treatment or all control) is bounded away from zero. We also require that the conditional expectation estimator $f$ and the true conditional expectation function $f^*$ be bounded in $L^q$ norm for $q > 4$.
    \begin{assumption}[Regularity conditions for GATE estimation in switchback experiments] \label{assm:sb_regularity}
        There exist constants $C, \zeta > 0$ such that the following regularity conditions are met for all $t \in [T]$
        \begin{align}
            & {\pi^*((t-m):t, \mathbf{d})} \geq \zeta, \text{ a.s.} & \forall \mathbf{d} \in \{ \mathbf{0}, \mathbf{1} \} \label{eq:sbsupport} \\
            & \norm{f(D_t, X_t, H_t)}_{L^{4+\delta(P)}} < C   \label{assm:sb_f_reg}\\
            & \norm{f^*(D_t, X_t, H_t)}_{L^{4+\delta(P)}} < C \label{assm:sb_fstar_reg}
        \end{align}
    \end{assumption}

    In \cref{assm:sb_rates}, we require that the time-average of the $L^2$ norm of $f-f^*$ is going to zero as $T \to \infty$.
    \begin{assumption}[Rate conditions for GATE estimation in switchback experiments] \label{assm:sb_rates}
        The following rate conditions are met:
        \begin{align}
            &T^{-1} \sum_{t=1}^T\norm{(f-f^*)(D_t, X_t, H_t)}_{L^2(P)} = o(1) \label{eq:rateobstreat}
        \end{align}
    \end{assumption}

    (Since treatment assignments are randomized by the experimenter, propensity scores are known. This means we do not need to impose rate conditions on the convergence of $f$ to $f^*$, in line with the implications of standard DML theorems for inference in experiments \citep{chernozhukov_doubledebiased_2018}.)

    \begin{theorem} \label{thm:switchback}
    For the model defined in \cref{sub:switchbackestimator}, and the estimand and estimator defined in \Cref{def:sb_psistar,def:sb_psi}, under \cref{assm:sb_regularity,assm:sb_rates}, then with probability no less than $1-\gamma$, it holds $$\sqrt{T}\hat \sigma^{-1}(\psi(W_{1:T}; \hat\eta) - \psi^*) \convindist N(0, 1),$$ where $\hat \sigma$ is as defined in \cref{eq:sigmasqhatdef}.
    \end{theorem}

    \ifproofsinbody
        \noindent \textit{Proof of \Cref{thm:switchback}.} 
    As in the proof of \cref{thm:ade}, we just need to verify \Cref{assm:regularity} and \Cref{assm:rates} and apply \cref{thm:main}. 
    \Cref{assm:regularity}\cref{item:twicegd} is trivially verified by the definition of $\psi$.
    \Cref{assm:regularity}\cref{item:neymanorth} is verified by \Cref{lem:switchbackneymanorth}.
    For \Cref{assm:rates}, \cref{assm:2.1} is verified by \cref{eq:sbsupport,assm:sb_f_reg,assm:sb_fstar_reg} and the definition of $\varphi$.
    \Cref{assm:2.2} is verified by \cref{lem:switchbackconvinnuisimpliesconvinest}.
    \Cref{assm:2.3} is verified by \cref{lem:switchbacksecondgd}.
\qed

\vspace{1em}

\Cref{lem:switchbackneymanorth} states that our estimator is Neyman orthogonal with respect to the nuisance realization set.
The proof resembles standard arguments of AIPW estimators, except that in our case the inverse-propensity is the replaced with the probabilities that the last $m$ observations where all assigned to treatment or control.

\begin{lemma}[Neyman orthogonality] \label{lem:switchbackneymanorth}
    For the model defined in \cref{sub:switchbackestimator}, and the estimand and estimator defined in \Cref{def:sb_psistar,def:sb_psi}, under \cref{assm:sb_regularity,assm:sb_rates}, then $\psi(W_{1:T}; \eta)$ is Neyman orthogonal with respect to $S_T$.
\end{lemma}

\noindent \textit{Proof of \Cref{lem:switchbackneymanorth}.} Notice: 
\begin{align*}
    &\frac{\partial }{\partial r}\E \big[ {\varphi(W_t; \eta^* + r(\eta - \eta^*))}\big] \big|_{r=0} \\
    &= \ex{(f-f^*)(1, X_t, H_t(\1)) - (f-f^*)(0, X_t, H_t(\0))} \\
    &\quad \paren{\frac{\indic{\dvec = \1}}{\pi^*({(t-m):t}, \1)} - \frac{\indic{\dvec = \0}}{\pi^*({(t-m):t}, \0)}}(\tilde Y_t + (f-f^*)(D_t, X_t, H_t) \big] \\
    &= \E \big[ (f-f^*)(1, X_t, H_t(\1)) \paren{1 - \frac{\indic{\dvec = \1}}{\pi^*({(t-m):t}, \1)}} \\
    &\quad - (f-f^*)(0, X_t, H_t(\0)) \paren{1 - \frac{\indic{\dvec = \0}}{\pi^*({(t-m):t}, \0)}}\big]
\end{align*}
where the first equality follows from \cref{eq:sbsupport,assm:sb_f_reg,assm:sb_fstar_reg} and applying the dominated convergence theorem. The second equality follows by the fact that $\ex{\tilde Y_t\; | \; D_{(t-m):t}} = 0$ and rearranging.
Now, using the facts that 
\begin{align*}
    &\expt{\indic{\dvec = \mathbf{1}} \; | \; \xvec } = \expt{\indic{\dvec = \mathbf{1}}} = \pi^*((t-m):t, \1), \;\; \text{and} \\
    &\expt{\indic{\dvec = \0} \; | \; \xvec} = \expt{\indic{\dvec = \0}} = \pi^*({(t-m):t, \0}),
\end{align*}
and iterated expectations, the Gateaux derivative is zero.
\qed

\vspace{1em}

\Cref{lem:switchbackconvinnuisimpliesconvinest} states the $L^2$ convergence of $\varphi(W_t;\eta)$, uniformly over the nuisance realization set $S_T$, to $\varphi(W_t; \eta^*$.
Similar to the previous section, verifying the result reduces to conditions on the convergence of the nuisances, which in this case is satisfied by the a requirement that $f$ converges to $f^*$ and treatment probabilities are known.

 \begin{lemma}[Consistency] \label{lem:switchbackconvinnuisimpliesconvinest}
    For the model defined in \cref{sub:switchbackestimator}, and the estimand and estimator defined in \Cref{def:sb_psistar,def:sb_psi}, under \cref{assm:sb_regularity,assm:sb_rates},
    then
    \begin{align*}
        \sup_{\eta \in S_T}T^{-1} \sum_{t=1}^T \norm{ \varphi(W_{t}; \eta) - \varphi(W_{t}; \eta^*)}_{L^2(P_T)} = o(1).
    \end{align*}
\end{lemma}

\noindent \textit{Proof of \Cref{lem:switchbackconvinnuisimpliesconvinest}} Notice
    \begin{align*}
        \varphi(W_{t}; \eta) - \varphi(W_{t}; \eta^*) &= (f-f^*)(1, X_t, H(\1)) - (f-f^*)(0, X_t, H(\0)) \\
        &\quad + \paren{\frac{\indic{\dvec = \1}}{\pi^*({(t-m):t}, \1)} - \frac{\indic{\dvec = \0}}{\pi^*({(t-m):t}, \0)}}(f-f^*)(D_t, X_t, H_t).
    \end{align*}
    Now, since, by \cref{eq:sbsupport} and \cref{eq:rateobstreat}, it holds
    \begin{align*}
        \norm{(f-f^*)(\mathbf{b}, X_t, H(\mathbf{b}))}_{L^2(P)} = o(1), && \forall \mathbf{b} \in \{ \0, \1 \}
    \end{align*}
    and by the fact that
    \begin{align*}
        \abs{\frac{\indic{\dvec = \1}}{\pi^*({(t-m):t}, \1)} - \frac{\indic{\dvec = \0}}{\pi^*({(t-m):t}, \0)}} \leq \max \curly{\pi^*((t-m):t, \1)^{-1}, \pi^*((t-m):t, \0)^{-1}} \leq \zeta^{-1}
    \end{align*}
    almost surely,
    the triangle inequality proves the result.
\qed

\vspace{1em}

\Cref{lem:switchbacksecondgd} states the condition that second-order terms on the path from $\eta$ in $S_T$ to $\eta^*$ do not dominate. 
In this case, the condition is easily verified by the fact that the setting is experimental and therefore treatment probabilities are known.

\begin{lemma}[Second-order condition] \label{lem:switchbacksecondgd}
    For the model defined in \cref{sub:switchbackestimator}, and the estimand and estimator defined in \Cref{def:sb_psistar,def:sb_psi}, under \cref{assm:sb_regularity,assm:sb_rates}, then
    \begin{align*}
        \sup_{r \in(0,1), \eta \in S_T}\abs{ \frac{\partial^{2}}{\partial r^{2} } \expt{ \psi(W_{1:T}; \eta^* + r(\eta - \eta^*))}} = o({T^{-1/2}}).
    \end{align*}
\end{lemma}

\noindent \textit{Proof of \Cref{lem:switchbacksecondgd}.} 
    The result holds trivially by  \cref{eq:sbsupport,assm:sb_f_reg,assm:sb_fstar_reg} and applying the dominated convergence theorem (to switch the derivative and expectation) and the fact that the second derivative is 0.
\qed

    \fi

     \subsection{Simulations and estimator comparison} \label{sub:sb_ni}

     We next validate the DML method for estimation of the GATE in switchback experiments.
     
\begin{figure}[t]
         \centering
         \ificml
\else
             \includegraphics[width=0.7\linewidth]{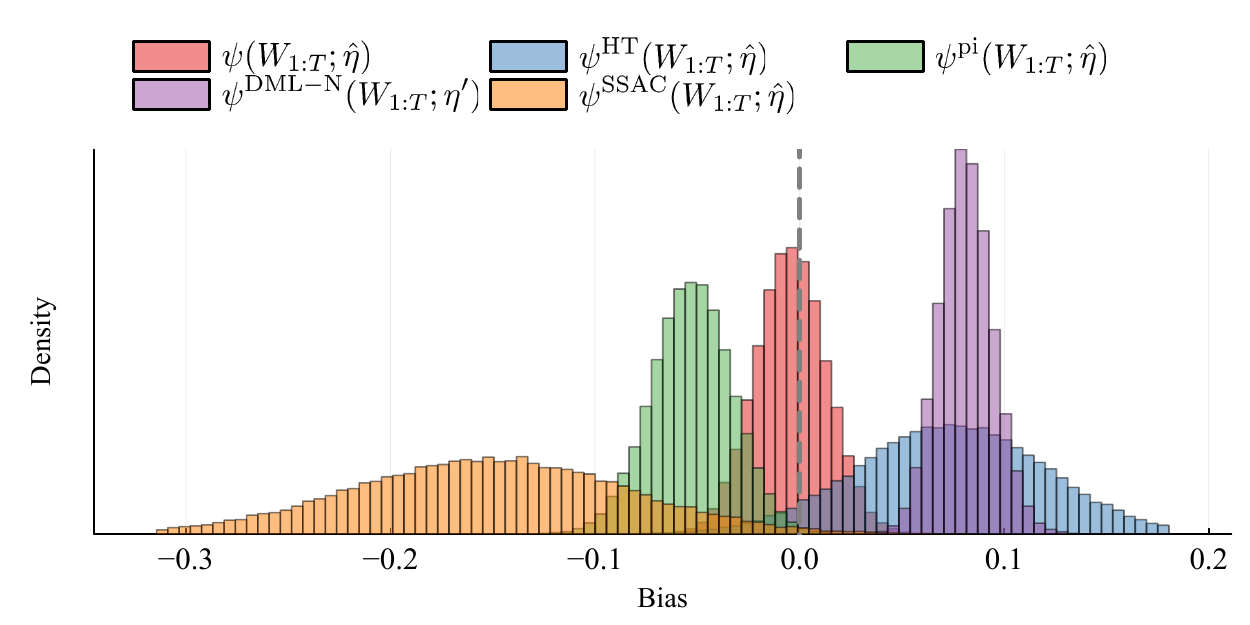}
         \fi
         
         \caption{Estimates of the global average treatment effect in switchback experiments with our double machine learning estimator versus naive estimators.}
         \label{fig:dmlsb-vs-naive}
     \end{figure}

\begin{figure}[t]
         \centering
         \ificml
\else
         \includegraphics[width=0.8\linewidth]{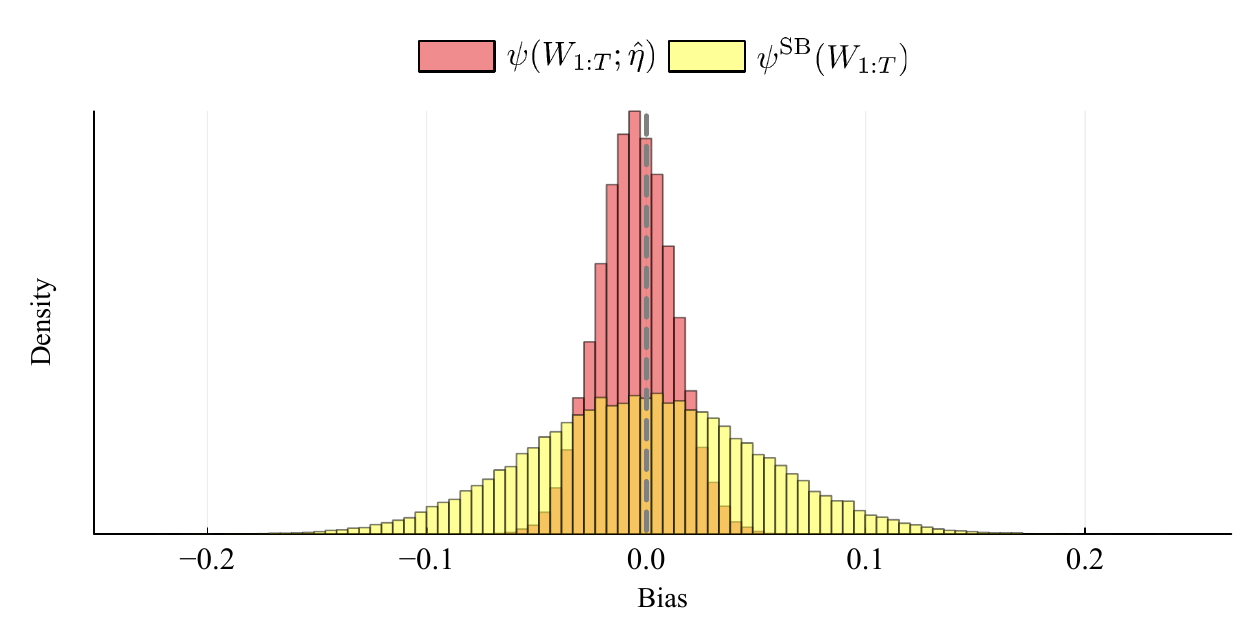}
         \fi

         \caption{Estimates of the GATE in switchback experiments with the double machine learning estimator versus the switchback Horvitz-Thompson estimator. The DML estimator has substantially lower variance.}
         \label{fig:dmlsb-vs-sb}
     \end{figure}

     \paragraph{Setting.} To guarantee $m$-dependence, the data generating process used for the plots in this section is different from that in \cref{sub:ade_ni}.
     In particular, the data generating process for $H_t$ in \cref{sub:ade_ni} allowed for dependence on prior shared states, which allows for long-range dependencies between observations over time. 
     In this section, $H_t$ is a function only of the last $m$ observations of treatments $D_{t-m:t-1}$ and covariates $X_{t-m:t-1}$.
     The setup is otherwise similar to that in \cref{sub:ade_ni}: $f^*$ is a smooth function of $D_t, X_t, H_t$ and we generate the predictor $\hat f$ using a random forest with default parameters.
     %
     %
     The switching length is chosen arbitrarily $\ell = 2m$. 
     Full details of the simulations are deferred to \cref{sec:simulationdetails}.
     We compare the performance of our estimator against naive estimators in \cref{fig:dmlsb-vs-naive} and against an unbiased estimator for treatment effects in switchback experiments in \cref{fig:dmlsb-vs-sb}.

    \paragraph{Treatment effect estimator comparisons.} In \cref{fig:dmlsb-vs-naive}, we compare our DML estimator to naive estimators.
    As in \cref{sub:ade_ni}, we compare our estimator to a naive Horvitz-Thompson estimator, a plug-in estimator and a naive DML estimator.
    Each estimator is defined the same way as in \cref{sub:ade_ni}, except that we plug in the known true propensity score $m^*$ instead of the estimated propensity score $\hat m$ since $m^*$ is known in an experiment.
    Additionally, we show the shared state as covariates estimator defined as
    \begin{align}
        \begin{aligned}
        \psi^{\mathrm{SSAC}}(W_{1:T}; \eta) &\defeq T^{-1} \sum_{t=1}^T  f(1, X_t, H_t) - f(0, X_t, H_t) \\&\quad\quad\quad\quad+ \paren{\frac{D_t}{ m(X_t)} - \frac{1 - D_t}{1 - m(X_t)}} \cdot (Y_t - f(D_t, X_t, H_t)).
        \end{aligned} \label{eq:ssac_te_estimator}
    \end{align}
    $\psi^{\mathrm{SSAC}}$ is the estimator obtained by treating the shared state as iid covariates.

    Note that the naive estimators are biased away from the true GATE, while the DML estimator is approximately centered at the true effect.
    The SSAC estimator in this section is biased, intuitively, because it \textit{controls for} the spillover effects between units, rather than \textit{including} them in the effect estimate: the global average treatment effect includes both the direct effect of treatment and the indirect effects of others' treatments.
    %
    %
    %
     
     In \cref{fig:dmlsb-vs-sb}, we compare our estimator with an unbiased switchback estimator proposed in \citet{bojinov_design_2022}, which is defined as
     \ificml
\else
     \begin{align}
         \psi^{\mathrm{SB}}(W_{1:T}) \defeq \sum_{t=1}^T \paren{\frac{\indic{\dvec = \1}}{\pi^*({(t-m):t}, \1)} - \frac{\indic{\dvec = \0}}{\pi^*({(t-m):t}, \0)}} Y_t .\label{eq:htsb}
     \end{align}
     \fi
     %
     %
     Note that both the DML and switchback HT estimator distributions are centered at the true effect, but the DML estimator has substantially lower variance.
     This is in accordance with the fact that some of the variation in outcomes is explained by the shared state and covariates (and is learned by expressive machine learning models), and the fact that the DML estimator uses all of the data while the switchback HT estimator drops the first $m$ observations after any switch.
     %
     %
     We leave more detailed exploration of the performance of the estimand (compared to (1) the Horvitz-Thompson-style switchback estimator and (2) across different choices of $\ell$) for future work. 

     \begin{figure}
         \centering
         \begin{subfigure}{0.49\textwidth}
             \centering
             \includegraphics[width=\linewidth]{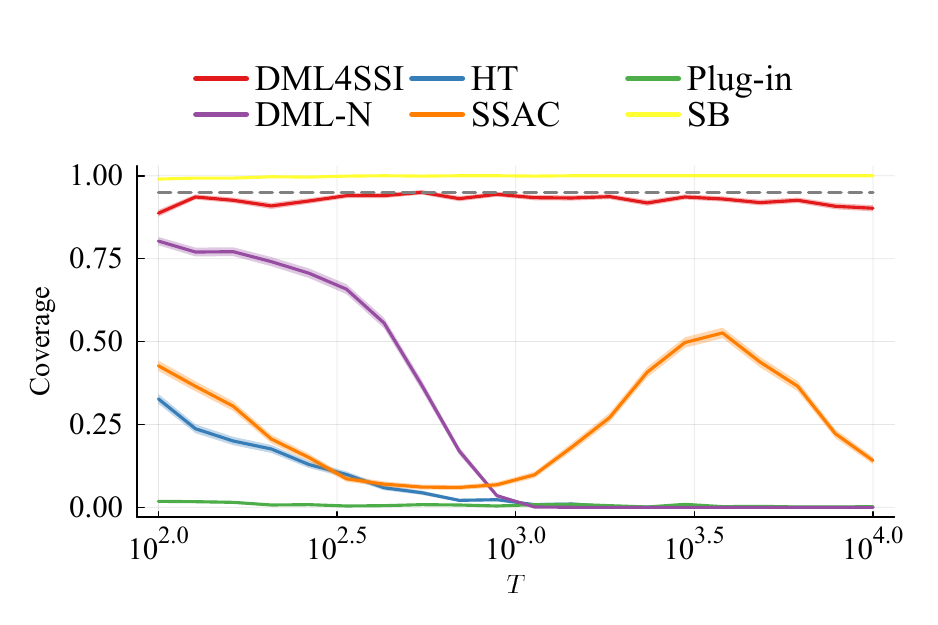}
             
         \end{subfigure}
         \begin{subfigure}{0.49\textwidth}
             \centering
             \includegraphics[width=\linewidth]{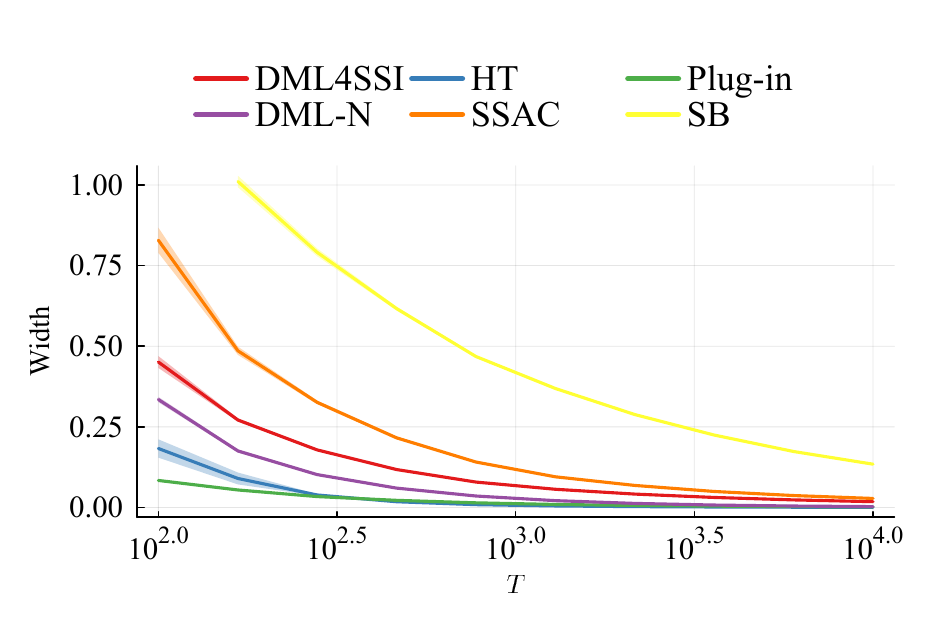}
             
         \end{subfigure}
         \caption{Coverage rates and widths of 95\% confidence intervals for the GATE constructed using our estimators (DML4SSI), switchback Horvitz-Thompson (HT) estimators, naive HT estimators, plug-in estimators and naive DML estimators. In the left plot, the coverage for confidence intervals constructed using the naive HT, plug-in and naive DML estimators overlap around 0. We plot standard errors of the simulations in lighter-colored ribbons around the coverage point estimates.}
         \label{fig:sb-coverage}
     \end{figure}

     \paragraph{Confidence interval comparisons.} We next explore the coverage of confidence intervals constructed from treatment effect and variance estimators.
     We use the same procedure to generate confidence interval coverage \cref{fig:sb-coverage} as we do for \cref{fig:coverage}: we run many simulations where we generate confidence intervals and compute the fraction of them that contain the true effect $\psi^*$.
     The definitions of the treatment effect and variance estimators for the naive HT, plug-in, naive DML estimators are the same as in \cref{sub:ade_ni}.
     We use our DML treatment effect estimator $\psi(W_{1:T}; \hat\eta)$ specified by \cref{def:sb_psi} and the consistent variance estimator from \cref{thm:main}.
     For the switchback HT estimators, we use the treatment effect estimator from \cref{eq:htsb} and the conservative variance estimator provided in \citet[Corollary 1]{bojinov_design_2022}. For the SSAC estimators, we use the treatment effect estimator from \cref{eq:ssac_te_estimator} and the naive variance estimator as defined in \cref{sub:ade_ni}.

     We observe that the DML4SSI confidence intervals are close to the target 95\% coverage, while the switchback HT estimator provides substantially greater than 95\% coverage.
     This is consistent with the fact that their variance estimator is conservative and will not be consistent in general.
     The mean confidence interval width constructed from the SB estimator (yellow) is between 3 and 7.5 times the width of that constructed from the DML4SSI estimator (red): put another way, in these simulations, the DML4SSI estimator requires many fewer samples to produce confidence intervals the same width as the SB estimator for a given time horizon $T$.
     Finally, we note that the naive estimators and plugin estimators all provide coverage near zero.
     This occurs as a result of the biases of each of these estimators for estimation of the treatment effect.


\section{Discussion} \label{sec:discussion}

    In this paper, we formally define shared-state interference and develop methods for efficient estimation of treatment effects using double machine learning.
    Shared-state interference occurs when spillovers in a system are channeled through a low-dimensional statistic like prices, inventory or information.
    Our motivating examples of shared-state interference are marketplaces and recommender systems, but there are many contexts where researchers and practitioners wish to measure treatment effects and where shared-state interference is present.
    Our method allows for the use of expressive machine learning estimation of nuisance parameters while achieving $\sqrt{T}$-rates.
    We also provide a consistent variance estimator.

    In future work, it would be valuable to explore a number of theoretical and empirical questions around shared-state interference.
    Theoretically, it would be interesting to explore what other estimands (e.g., the average indirect effect as in \citet{munro_treatment_2023} or time-discounting as in \citet{farias_correcting_2023}) or structural models allowing for inference using our \cref{thm:main}.
    It would also be valuable to characterize alternate forms of interference that have a similar structure: for example, in some situations, a shared state may influence treatment assignments (which we assume does not occur in our structural models). 
    Empirically, it would be interesting to apply our methods to markets or information systems that exhibit shared-state interference.
    We hope that this work inspires further investigations of tractable interference structures in semiparametric inference.


\ificml
\else
\pagebreak
\fi

\bibliographystyle{abbrvnat}
\bibliography{main}

\appendix
\onecolumn

    \section{Background} \label{sec:background}

    \paragraph{Additional notation.}
    For a vector-valued random variable $V$, when we write $V = O_p(\cdot)$, we mean that the relation holds element-wise.
    For a sample space $\Omega$, let $\cB(\Omega)$ be the corresponding Borel set.
    
    \paragraph{Geometrically ergodic Markov chains.} 
    A geometrically ergodic Markov chain is a Harris ergodic chain that satisfies a geometric mixing rate property. 
    A Harris ergodic chain is aperiodic, $\mu$-irreducible where $\mu$ is a ``maximal'' irreducibility measure (see, e.g., \cite[Theorem 4.0.1]{meyn_markov_2009}) and positive Harris recurrent. 
    We define each of these properties next.
    
    First, we review aperiodicity for our setting. 
    The period of the chain $k$ is defined in our context as the largest sequence of disjoint measurable sets $A_0, \dots, A_{k-1}$ such that the chain cycles between the sets with probability one: i.e., that $P(W_t \in A_{(i+1) \% k} \; | \; W_{t-1} \in A_{i}) = 1$ for all $i \in \{0, \dots, k-1\}$.
    An aperiodic chain is one where the period is 1.
    A chain is $\mu$\textit{-irreducible} for some measure $\mu$, if, for all measurable $A \subseteq \cW$ such that $\mu(A) > 0$, it holds for all $w \in A$ and $i \in [T]$
    \begin{align*}
         P(\exists \, t \in \N \; : \; W_{t+i} \in A \; | \; W_i = w) > 0.
    \end{align*}
    This definition of irreducibility is a generalization of irreducibility in finite-state Markov chains.
    \textit{Harris recurrent} means that for all $\mu$-measurable $A \subseteq \cW$ with $\mu(A) > 0$ and $w \in A$, it holds,
    \begin{align*}
        P( \{ W_{t} \}_{t=1}^\infty \in A \; \mathrm{i.o.}  \; |\;  W_1 = w ) = 1,
    \end{align*}
    where i.o. stands for infinitely often:
    \begin{align*}
       \{ \{ W_{t} \}_{t=1}^\infty \in A \; \mathrm{i.o.} \} \defeq \bigcap_{s=1}^{\infty} \bigcup_{t=s}^\infty \{ W_t \in A \}.
    \end{align*}
    Harris recurrence is a generalization of recurrence in finite-state Markov chains.
    A chain is \textit{positive} if there exists a measure $\tau$ such that, for all $\tau$-measurable $A$, 
    \mr{You use $P$ for a placeholder measure here, and $P$ is also used elsewhere in the paper. Make sure it isn't overloaded.}\ch{Good catch. Fixed.}
    \begin{align*}
        \tau(A) = \int_{\cW} K(w, A) \tau(dw),
    \end{align*}
    i.e., that there exists an invariant measure.

    \textit{Detailed balance} is defined in \cref{def:detailedbalance}. 
    The property means, for continuous state spaces, that the steady state probability density of each state times the probability density of the transition to another state is equal to the steady state probability density times the density of transition to the first state.
    
    A Harris ergodic Markov chain is \textit{geometrically ergodic} if there exists a function $M \; : \; \cW \to \R_{>0}$ and constant $\theta \in (0, 1)$ such that, for all $w \in \cW$,
    \begin{align*}
        \tvnorm{P^T(w, \cdot) - P_{\infty}(\cdot)} \leq M(w) \theta^T,
    \end{align*}
    where $\tvnorm{\cdot}$ is the total variation distance norm defined for a signed measure $\mu$ as
    \begin{align*}
       \tvnorm{\mu} = \sup_{A \in \cB(\cW)} \mu(A) - \inf_{A \in \cB(\cW)} \mu(A).
    \end{align*}

    \textbf{Double machine learning (DML).} Double machine learning is a meta-algorithm for semiparametric inference, introduced and analyzed in \citet{chernozhukov_doubledebiased_2018}.
    It allows for the use of expressive machine learning models even when those models may not converge to a true data generating process at parametric rates.
    The algorithm consists of two conceptual components: sample splitting and a bias correction in the estimator.
    The first component is the construction of nuisance function approximators from data that is independent of the observed data it will be applied to, typically achieved by sample splitting.
    The nuisance estimators can be constructed using expressive machine learning methods that may produce biased predictions and converge at slower-than-parametric rates.
    The second component is the use of a bias correction term added to a plug-in estimator which ensures that errors in the nuisance function approximators depend only on second-order terms.
    We use the following definitions of the Gateaux derivative and Neyman orthogonality in our results.
    
    \begin{definition}[Gateaux derivative] \label{def:gateaux} For a convex set of functions $\cS$ and function $\eta \in \cS$, the \textit{Gateaux derivative} of a map $f \; : \; \cS \to \R$ with respect to $\eta, \eta^* \in S$ is defined as the quantity
    \begin{align*}
        \frac{\partial}{\partial r} {f(\eta^* + r(\eta - \eta^*))}, \quad r \in [0,1].
    \end{align*}
    \end{definition}
    \begin{definition}[Neyman orthogonality] \label{def:neymanorth}
        For a distribution $P$ on data $W$, the function $\psi$ is said to be \textit{Neyman orthogonal with respect to $S$} for $\eta^* \in \cS$ and all $\eta \in \cS$ the Gateaux derivative of $\eta \mapsto \E_P[\psi(W; \eta)]$ with respect to $\eta, \eta^*$ vanishes at $r=0$:
    \begin{align*}
        \frac{\partial}{\partial r} {f(\eta^* + r(\eta - \eta^*))} \bigg|_{r=0} = 0.
    \end{align*}
    \end{definition}

\section{Additional Related Work} \label{sec:morerelatedwork}

Here we provide a detailed comparison of several prior works that are particularly closely related to our paper.

\textit{Comparison to \citet{chernozhukov_doubledebiased_2018}.} We prove a theorem analogous to their Theorem 3.1 for the case of dependent data.
We make two changes to their setup. 
The first is that we will not assume each individual's treatment assignment, covariates and outcomes $(D, X, Y)$ are drawn iid from a joint distribution. 
Instead, we will assume that covariates and treatment assignments are drawn iid, that units arrive sequentially, and that outcomes are a function of covariates and a shared state possibly depending on the treatment assignments, covariates and outcomes of previous arrivals.
Second, we assume, in the style of \citet{angelopoulos_prediction-powered_2023}, there is access to an auxilliary sample on which to train the nuisance parameters.
We will assume that the size of the auxilliary sample scales appropriately so that the nuisance functions converge in $L^2(P)$ at appropriate rates and are independent of the data.
In \citet{chernozhukov_doubledebiased_2018}, independence is achieved through sample splitting of the iid data.
In our context, where outcomes are not drawn iid, the auxiliary training data serves the role that sample splitting does in the double ML for iid data.
Notationally, our score function $\psi$ is defined differently than theirs: ours must be a function of all of the data $W_{1:T}$ whereas theirs may be a function of a single draw $W$ from the iid data distribution.

\textit{Comparison to \citet{emmenegger_treatment_2023}.} As in their main results, we derive efficient inference results for the average direct effect. In their appendix, they also provide results that are analogous to ours for estimation of the global average treatment effect. 
Our model can be written as a network interference model by constructing a network where each node affects the potential outcomes of future nodes.
However, they prove results about network interference under a sparsity assumption, and our network interference model is not sparse (since there would be a directed edge from each node $t$ to each future node $s > t$).
Instead, our framework makes the interference structure tractable by mediating interference through shared states.
Thus, our work allows for potentially long-range dependencies between observations that are disallowed by the sparsity assumptions in the network interference literature.
Additionally, our model allows for dependencies between units' \textit{outcomes} and future units' outcomes, which is disallowed in their model (and the vast majority of work in the network interference literature).

\textit{Comparison to \citet{ballinari_semiparametric_2024}.} Their DML results, like ours are about a time series model where units are observed in a sequence over time. They instantiate a structural model similar to our approach in \cref{sec:ade,sec:sb}, and their efficient influence results, like ours, go through mixing conditions. The estimand in their structural model is a impulse response function, which measures the effect of an intervention at time $t$ on outcomes at time $t+h$ for some known $h$, whereas ours is the average direct effect and global average treatment effect.
Our structural models are also different, since we require all interference across time to occur through shared-state variables, whereas they allow for generic mixing across time. 

\textit{Comparison to \citet{munro_causal_2024}.} This paper considers semiparametric estimation of the global average treatment effect in settings where outcomes are determined by a centralized mechanism like an auction, rather than a context- and mechanism-agnostic setting like outs. Informally, their key assumptions require that the mechanism allocates goods according to a known function of individuals' bids and market-clearing cut-offs. They also require individuals' ``bids'' to the mechanism obey the stable unit treatment value assumption (SUTVA). Their work establishes estimation of the global average treatment effect in observational settings, whereas we provide methods for estimating the average direct effect in observational settings and the global average treatment effect in experimental settings. Their model considers simultaneous arrivals of units to the market, whereas ours considers sequential arrivals.

\textit{Comparison to \citet{zhan_estimating_2024}.} We prove a theorem analogous to their main theorem, except that our results go through Markov chains and theirs go through martingales. We also do not make functional form assumptions about the relationship between outcomes and the shared state, which in their case is the set of recommendations chosen by the platform (they use a discrete choice model). We also do not require that the shared state be computed by a neural network, as they do.

\textit{Connections to the network interference literature.} 
    Our model can be interpreted as a network with a particular structure (where each unit affects the outcomes of future units), but our model allows for dependencies between outcomes of each unit and the outcomes of subsequent units.
    Typically, in the network interference literature, all dependencies are channeled through treatment assignments, rather than outcomes, since outcomes in their models are typically not sequentially determined (see, e.g. \citet{aronow_spillover_2020} for an overview).
    Additionally, methods in network interference often make a sparsity assumption on the interference graph (like those in \citet{emmenegger_treatment_2023}), whereas our interference network can be dense.
    However, since correlation between units dissipates over time, it would be interesting to explore whether models of approximate neighborhood interference \citep{leung_causal_2022} or approximate local interference \citep{chin_central_2019}. 
    


\ifproofsinbody
\else
    \section{Further details and proofs for \cref{sec:dml4ssi}} \label{sec:app_dml4ssi}
    \mr{In each of these sections with proofs, consider opening with a paragraph describing how the section is organized.}

    \mr{High-level comment on the technical sections: it reads as just a proof of correctness. This is a fine choice to make. But to the extent that you want to make it easier for people to build on what you're doing, consider the perspective of a 2nd-year PhD student who wants to work in this area. Can you help them pick up some intuition for how to think about the proofs, what the key steps are, etc. The organization suggestion above is an example of that, but throughout the proofs you can also draw attention and add explanations for things you think are particularly important/reusable.}
    \ch{I agree with this goal. I added paragraphs and a few sentences before each lemma. Is this close to what you have in mind?}

    In this section, we prove our main theorem in \cref{sec:dml4ssi}.
    At a high level, the two parts of the theorem are $\sqrt{T}$-asymptotic normality of the treatment effect estimator and consistency of the variance estimator.
    Here we overview the proofs of each of these parts.

    For $\sqrt{T}$-asymptotic normality of the treatment effect estimator, we make the simple observation that the difference between the treatment effect estimator $\psi(W_{1:T}; \hat\eta)$ and the estimand $\psi^*$ can be broken down into the sum of three differences.
    (This is an observation made in \citet[page 19, equation 10]{kennedy_semiparametric_2023} and is standard in proofs of DML theorems.)
    The three components in the sum are: (1) The difference between an estimator where the true nuisances are known (sometimes called an oracle estimator) and the estimand (\cref{eq:oracle_diff}), (2) the deviation from its mean of the difference between the oracle estimator and our DML estimator where the nuisances are not known (\cref{eq:empiricalproc}), and (3) the expected difference between the oracle and plugin estimators.
    Then, the first term is shown to be $\sqrt{T}$-asymptotically normal in \cref{lem:score}, and the second two terms can be shown to go to zero in probability at faster than $\sqrt{T}$-rates in \cref{lem:empiricalproc,lem:neymanorth}, respectively.
    Together, this implies $\psi(W_{1:T}; \hat\eta )$ approaches its mean at $\sqrt{T}$ rates and is asymptotically normal.
    The consistency of the variance estimator is given by \cref{lem:var}.
    We describe the high-level intuition for each of the lemmas before their proofs.

    \vspace{1em}


    \section{Further details and proofs for \cref{sec:ade}}


    In this section, we prove our results for \cref{sec:ade}.
    The result follows directly from \cref{thm:main} once we have verified the assumptions necessary for the theorem to apply.
    Thus, our main task is verifying these assumptions.

    There are three parts of the assumptions in \cref{sec:dml4ssi} that are not trivially verified by analogous assumptions in \cref{sec:ade}.
    These are: the Neyman orthogonality of $\psi$, $L^2$ convergence of $\varphi(W_t, \eta)$ to $\varphi(W_t, \eta^*)$ on average over $t$, and a second-order condition stating that the second-order misestimation of $\eta^*$ does not dominate.
    Each of these is stated in a separate lemma, and we provide high-level descriptions of each of them before they are stated and proved.

    \vspace{1em}

    \section{Further details and proofs for \cref{sec:sb}}


    In this section, we provide proofs for our results in \cref{sec:sb}.
    The proof of \cref{thm:switchback} mirrors that of \cref{thm:ade}: we just need to verify the assumptions necessary for applying \cref{thm:main}.
    This again reduces to verifying a Neyman orthogonality condition, a consistency condition and a second-order Gateaux derivative condition.

    \vspace{1em}

\fi

\section{Additional simulation details} \label{sec:simulationdetails}

\subsection{Simulations for \cref{sec:ade}} \label{sub:supsim}
    For each simulation, we set $T=$ 1,000 and $d_X = 10$. 
    For all $i \in 1,\dots,d_X$, we let $X_{t,i} \sim N(1, 1)$ iid.
    $D_t \sim \Ber(\min \{\max\{\zeta, X_{t,1} \} , 1- \zeta\})$ and $\zeta = 0.1$.
    The outcome function is defined as 
    \begin{align*}
        &Y_t(D_t, H_t) \defeq \sin (2 \pi X_{t,1}) + 2D_t + 2 H_t D_t - 1 + \tilde Y_t, \\
        &H_t(W_{t-1}) \defeq 0.75 (H_{t-1} - 1) + 1 + \tilde H_t.
    \end{align*}
    where $\tilde Y_t \sim \cN(0, 1/10), \tilde H_t \sim \cN(0, 1)$ iid.
    For \cref{fig:ade}, we run 50,000 simulations.
    In \cref{fig:coverage}, we run 1,000 simulations for each time horizon $T$ and plot $10$ values of $T$ evenly divided on a logarithmic scale from 100 to 10,000.

    In \cref{fig:coverage}, for each time horizon $T$, we compute 1,000 simulations. 
    We also compute the empirical standard error for each estimate in ribbons around the point estimates by computing the empirical standard deviation of the point estimates across simulations and dividing by the square root of the number of simulations.

\subsection{Simulations for \cref{sec:sb}}

    We let $d_X = 1$ and $X_t \sim \Unif([0,1])$. Also, we let
    $D_t \sim \Ber(\min \{\max\{\zeta, X_{t,1} \} , 1- \zeta\})$ where $\zeta = 0.1$.
    We let $H_t$ be the last $m$ observations $X_{t-m:t-1}, D_{t-m:t-1}$ where $m = 5$. We let
    \begin{align*}
        &Y_t(D_t, H_t) \defeq \sin (2 \pi X_{t,1}) + 2D_t + 2 \sum_{i \in \floor{m/6}} e^{-H_{t,6i}/3} - 1 + \tilde Y_t,
    \end{align*}
    where $\tilde Y_t \sim N(0, 1/10)$ iid.
    We set the switching period to $2m$.

    In our simulations, we let $T =$ 1,000. In \cref{fig:dmlsb-vs-sb,fig:dmlsb-vs-naive}, we run each simulation 50,000 times.
    In \cref{fig:sb-coverage}, for each time horizon $T$, we run 1,000 simulations.

\end{document}